\documentclass{article}

\usepackage{microtype}
\usepackage{graphicx}
\usepackage{cuted}
\usepackage{subfigure}
\usepackage{booktabs} 

\usepackage{hyperref}

\usepackage[accepted]{icml2024}

\usepackage{amsmath}
\usepackage{amssymb}
\usepackage{mathtools}
\usepackage{amsthm}
\usepackage{graphicx}

\usepackage[capitalize,noabbrev]{cleveref}

\theoremstyle{plain}

\theoremstyle{definition}

\theoremstyle{remark}

\usepackage[textsize=tiny]{todonotes}

\icmltitlerunning{BiLoRA: A Bi-level Optimization Framework for Overfitting-Resilient Low-Rank Adaptation}

\begin{document}

\twocolumn[
\icmltitle{BiLoRA: A Bi-level Optimization Framework for Overfitting-Resilient Low-Rank Adaptation of Large Pre-trained  Models}



\icmlsetsymbol{equal}{*}

\begin{icmlauthorlist}
\icmlauthor{Rushi Qiang}{equal,ttt}
\icmlauthor{Ruiyi Zhang}{equal,uuu}
\icmlauthor{Pengtao Xie}{uuu}
\end{icmlauthorlist}

\icmlaffiliation{ttt}{Tsinghua University}
\icmlaffiliation{uuu}{UC San Diego}

\icmlcorrespondingauthor{Pengtao Xie}{p1xie@ucsd.edu}


\vskip 0.3in
]



\printAffiliationsAndNotice{\icmlEqualContribution} 

\begin{abstract}

Low-rank adaptation (LoRA) is a popular method for fine-tuning large-scale pre-trained models in downstream tasks by learning low-rank incremental matrices. Though LoRA and its variants effectively reduce the number of trainable  parameters  compared to full fine-tuning methods, they often overfit training data, resulting in sub-optimal generalization on test data. To address this problem, 
 we introduce BiLoRA, an overfitting-alleviating  fine-tuning approach based on bi-level optimization (BLO).  
BiLoRA employs pseudo singular value decomposition to parameterize low-rank incremental matrices and  splits the training of pseudo singular vectors and values across two different subsets of training data. This division, embedded within separate levels of the BLO framework, mitigates the risk of overfitting to a single dataset. Tested on ten datasets covering natural language understanding and generation tasks and applied to various well-known large pre-trained models, BiLoRA significantly outperforms LoRA methods and other  fine-tuning approaches, with similar amounts of trainable parameters.   

\end{abstract}

\section{Introduction}
\label{introduction}

\begin{figure*}[t]
	
	\begin{minipage}{0.33\linewidth}
		\vspace{3pt}
		\centerline{\includegraphics[width=\textwidth]{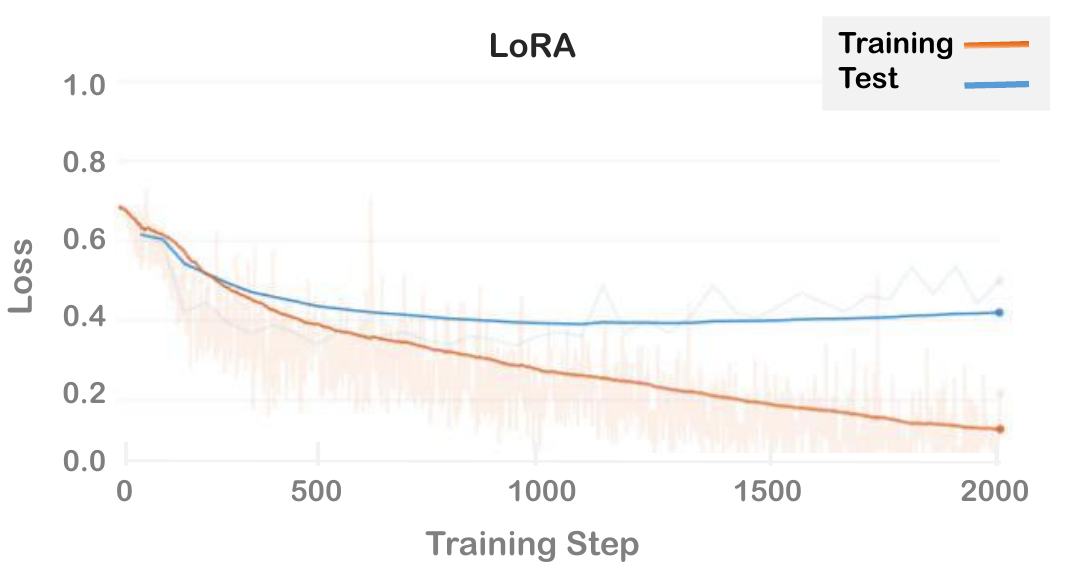}}
	\end{minipage}
	\begin{minipage}{0.33\linewidth}
		\vspace{3pt}
		\centerline{\includegraphics[width=\textwidth]{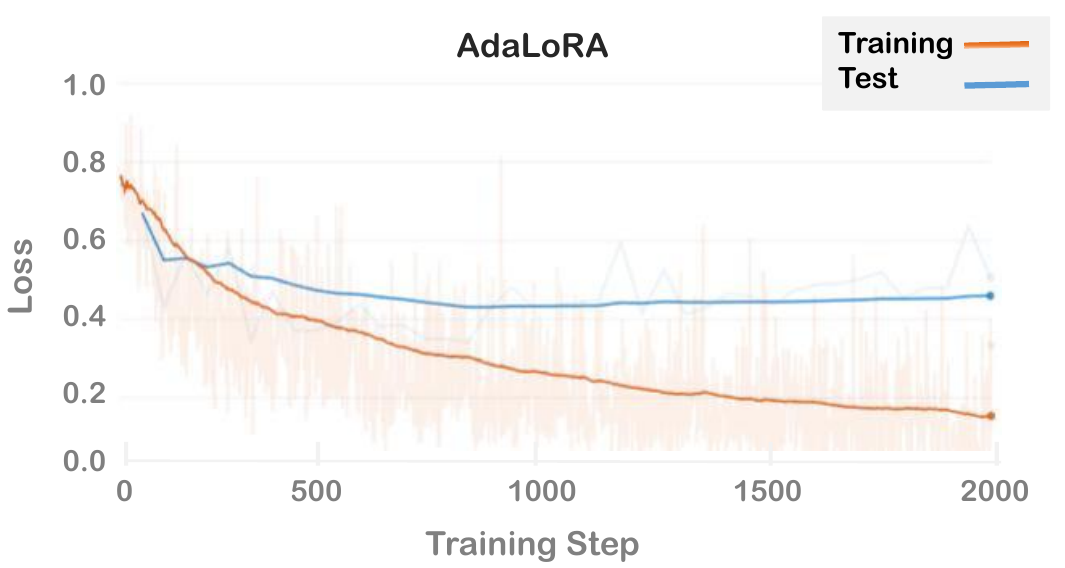}}
	 
	\end{minipage}
	\begin{minipage}{0.33\linewidth}
		\vspace{3pt}
		\centerline{\includegraphics[width=\textwidth]{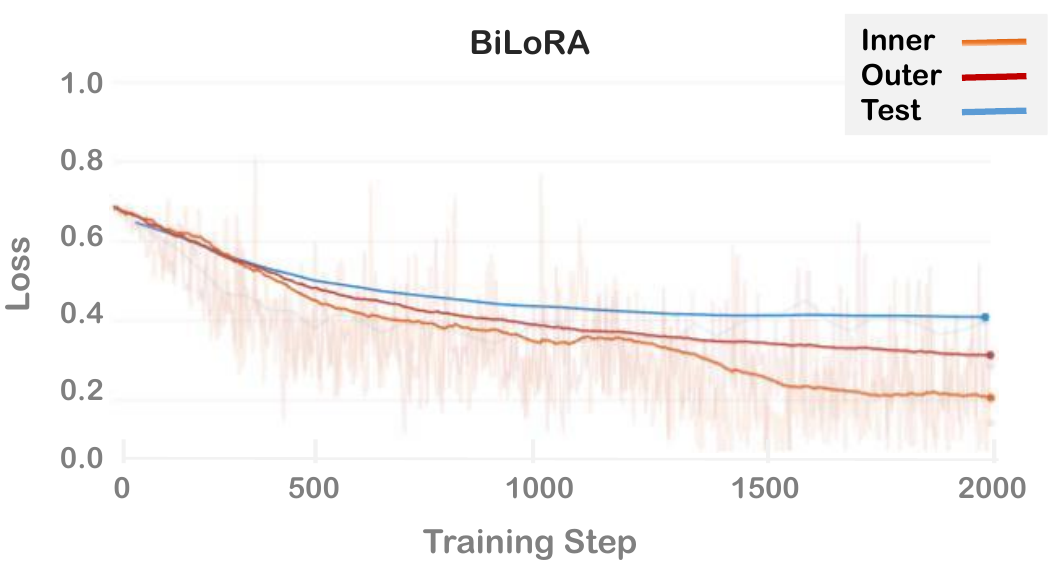}}
	 
	\end{minipage}
 
	\caption{Loss curves  on CoLA training  and test  datasets. 
The model  being fine-tuned is RoBERTa.}
\vspace{-0.4cm}
	\label{overfitting}
\end{figure*}

Large language models (LLMs) have demonstrated remarkable capabilities in a variety of natural language processing tasks~\citep{devlin2018bert, he2020deberta, radford2019language, brown2020language}. The typical approach to applying LLMs in real-world applications involves two stages: initial pre-training on extensive datasets, followed by fine-tuning on specific downstream tasks. However, with the increasing size of LLMs, full fine-tuning~\citep{qiu2020pre}, which involves updating all model parameters, incurs substantial computation costs.  Moreover, the extensive parameter count in these pre-trained models can lead to a high risk of overfitting during fine-tuning~\citep{karimi2021compacter}. To address these challenges, various Parameter-Efficient Fine-Tuning (PEFT) methods~\citep{houlsby2019parameter, ding2023parameter, mao2021unipelt} have been developed, which  aim to minimize the number of parameters that require fine-tuning, while still preserving the models' performance.


Low-Rank Adaptation (LoRA) \citep{hu2021lora} is a prominent PEFT method. 
It introduces low-rank update matrices (LRUMs) to pre-trained weight matrices. These LRUMs are compactly represented as the product of two much smaller matrices. During the fine-tuning process, only the LRUMs are adjusted, while the original pre-trained weights remain unchanged.   LoRA and its variants, such as AdaLoRA~\citep{zhang2023adaptive}, effectively reduce the number of trainable parameters  compared to traditional full fine-tuning. However, our comprehensive experiments indicate that these methods are still prone to significant overfitting. Figure~\ref{overfitting} provides illustrative examples of this trend. As fine-tuning progresses, the disparity between training and testing losses in both LoRA and AdaLoRA  becomes more pronounced. Beyond a certain number of iterations, we observe an increase in test losses alongside a continued decrease in training losses, clearly indicating a tendency for LoRA and AdaLoRA to overfit to the training data.

To overcome the limitations of traditional LoRA methods, we introduce BiLoRA, a novel fine-tuning approach designed to prevent overfitting through bi-level optimization (BLO). Bi-level optimization~\cite{sinha2017review} involves two nested optimization problems: the optimal variables at the lower level serve as inputs for the upper level's objective function, while the non-optimal variables at the upper level serve as inputs for the lower level's objective function. 
In BiLoRA, we  parameterize   each low-rank update matrix as $\Delta W = P \Lambda Q$,  akin to singular value decomposition. To approximate $\Lambda$ as a singular value matrix, we apply regularization to ensure the orthogonality of $P$ and $Q$.
 At the lower level of our formulation, we  train the matrices $\{P,Q\}$ by minimizing a fine-tuning loss on a  subset $S$ of the training dataset $D$. During this phase, $\Lambda$ is held constant. The resulting optimally learned matrices, $\{P^*(\Lambda),Q^*(\Lambda)\}$, are directly dependent on $\Lambda$. 
Subsequently, at the upper level, we evaluate $\{P^*(\Lambda),Q^*(\Lambda)\}$ using the remaining part of the dataset, $D\backslash S$. The validation loss, a function of $\Lambda$, guides the learning process for $\Lambda$ by minimizing this loss. 
By  partitioning the learning processes of $\{P,Q\}$ and $\Lambda$ across distinct  subsets of data and different levels of optimization problems, BiLoRA effectively mitigates overfitting to a specific dataset.

BiLoRA's mechanism of combating overfitting is inspired by the well-established practice of Differentiable Architecture Search (DARTS)~\cite{liu2018darts}. Typically, weight parameters of candidate operations (e.g., convolution, pooling) in a search space are trained using a training dataset, while the architecture - characterized by learnable scores that determine the selection of these operations for the final model - is learned using  a separate validation set. This approach prevents overfitting to the training data. If the architecture was also learned using the training set, it would likely result in an overly complex model, incorporating all possible candidate operations to fit the training data closely. However, such a model would exhibit poor generalization when applied to test data, as it would be specifically tailored to the training dataset's characteristics. 
In the LoRA framework, pseudo singular values can be conceptualized as an `architecture', while pseudo singular vectors are akin to candidate operations. This analogy becomes clearer when we write the SVD form of the update matrix, which is $\Delta W = P \Lambda Q$, equivalently  as \(\Delta W = \sum_{i=1}^r \Lambda_{ii} P_i Q_i^\top\), which represents the weighted sum of \(r\) rank-1 matrices. Each matrix is formed by a pair of left and right singular vectors, \(P_i\) and \(Q_i\), and is weighted by the  singular value \(\Lambda_{ii}\). In this context, each rank-1 matrix \(P_i Q_i^\top\) can be viewed as a `candidate operation'. The corresponding singular value \(\Lambda_{ii}\), which adjusts the weighting of the operation \(P_i Q_i^\top\) in the summation, functions as an `architecture' variable. 
Overfitting occurs when both the `architecture' \(\Lambda\) and `candidate operations' \(\{P,Q\}\) are simultaneously optimized by minimizing a loss function on a single dataset, as is the case with existing LoRA methods. 
In contrast, our BiLoRA approach aligns with the proper implementation of DARTS. It  optimizes the `architecture' \(\Lambda\) using a `validation set', which is a subset of the training data, while the `candidate operations' \(\{P,Q\}\) are trained on a different subset of the training data. Thereby, our method is more resilient to overfitting.


Our key contributions are outlined as follows:

\begin{itemize}
    \item We introduce a novel bi-level optimization approach to mitigate overfitting in LoRA and its variants. Unlike traditional methods that train an entire incremental matrix on a single dataset, 
    our approach divides the learning of distinct parameter subsets across different sub-datasets and different levels of optimization problems which are closely interconnected.    This strategy effectively reduces overfitting to any single dataset.
    \item Our method's efficacy is validated across ten datasets in both natural language understanding and generation tasks, utilizing major pre-trained models such as RoBERTa, DeBERTa, and GPT2. When compared to LoRA, AdaLoRA, and other prevalent finetuning  methods, our approach demonstrates superior performance while maintaining a comparable number of trainable parameters.
\end{itemize}

\section{Related Work}

\label{gen_inst}

\textbf{Low-Rank Adaptation.}
\citet{li2018measuring} and \citet{aghajanyan2020intrinsic} demonstrate that widely-used pre-trained models possess a very low intrinsic dimension and it is possible to achieve comparable fine-tuning performance by utilizing a reparameterization with  reduced dimensionality. This inspires low-rank adaptation (LoRA) to be introduced for fine-tuning LLMs.
LoRA introduces incremental updates to frozen pre-trained weights as low-rank matrices \citep{hu2021lora}. By parameterizing an update  matrix as the product of two low-rank matrices, LoRA greatly reduces trainable parameters while maintaining or even improving the performance over full fine-tuning. Multiple methods have been proposed to improve the time/memory efficiency and performance of LoRA. 
DyLoRA \citep{valipour2022dylora} optimizes low-rank updates  with multiple ranks by sorting  learned representations dynamically during training. QLoRA \citep{dettmers2023qlora} introduces multiple strategies to reduce memory footprint for LoRA, lowering the memory barrier for fine-tuning  LLMs. LoraHub \citep{huang2023lorahub} is designed to facilitate the efficient combination of LoRA modules trained on various tasks using only a few examples from a new task.  AdaLoRA \citep{zhang2023adaptive} allocates the parameter budget adaptively according to the importance of modules to improve the fine-tuning performance in specific budget settings. It parameterizes the incremental updates in the form of singular value decomposition and iteratively prunes singular values in correspondence to their importance scores during training. 
Different from these existing methods which train all the parameters in incremental updates on a single training dataset and therefore often lead to overfitting, our method (based on the SVD reparameterization of incremental updates)  separately trains  singular values and singular vectors in two different optimization levels, which effectively alleviates the risk of overfitting to a single dataset. 

\textbf{Bi-level Optimization (BLO).} 
BLO has gained much attention for formulating various machine learning methods  including meta-learning \citep{finn2017model, rajeswaran2019meta}, hyperparameter optimization \citep{franceschi2017forward, lorraine2020optimizing}, neural architecture search \citep{liu2018darts,zhang2021idarts}, reinforcement learning \citep{rajeswaran2020game}, to name a few. In addition to applying BLO to various machine learning problems, various algorithms have been proposed to address this specific form of optimization problem, including  zeroth-order methods like Bayesian optimization \citep{cui2019new}, first-order algorithms based on hypergradients  \citep{pearlmutter2008reverse, lorraine2020optimizing}, etc. Gradient-based BLO is efficient for scaling up to high-dimensional problems with a large number of trainable parameters. We expand the application scenarios of gradient-based BLO and build an efficient training framework  to improve the generalization performance of LoRA.

\section{Methods}
\label{headings}

\begin{figure}[t]
\centering
\includegraphics[width=0.35\textwidth]{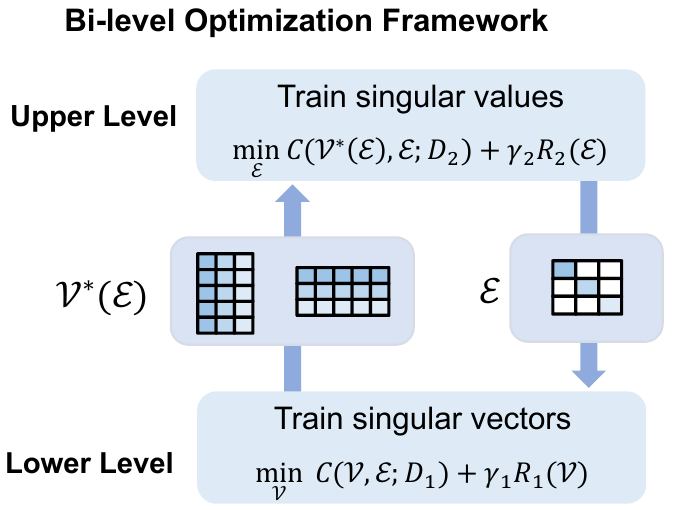}
\vspace{-0.3cm}
\caption{The proposed BiLoRA  method.}
\label{fig:overview}
\vspace{-0.5cm}
\end{figure}

We propose BiLoRA (Figure~\ref{fig:overview}),  a novel LoRA-style fine-tuning framework based on bi-level optimization. Similar to AdaLoRA, incremental matrices in our method are parameterized in a pseudo SVD form with learnable pseudo singular vectors $\mathcal{V}$ and pseudo singular values $\mathcal{E}$. We split the training dataset into two non-overlapping subsets $D_1$ and $D_2$. 
In the lower level, we train $\mathcal{V}$ on $D_1$ while fixing $\mathcal{E}$. The optimal solution $\mathcal{V}^*(\mathcal{E})$ (which is a functional of $\mathcal{E}$) is fed into the upper level. In the upper level, we train  $\mathcal{E}$ on the dataset $D_2$. The updated $\mathcal{E}$ is fed into the lower level. The two levels of optimization problems are solved iteratively until convergence.  

\subsection{Parameterization of Low-Rank Incremental Matrices}
\label{sec:para}

Following~\citep{zhang2023adaptive}, we parameterize a low-rank incremental  matrix $\Delta W$ as $\Delta W = P \Lambda Q$  which mimics SVD. The diagonal matrix $\Lambda$ contains \textit{pseudo singular values} and the approximately orthogonal matrices $P$ and $Q$ represent \textit{pseudo left/right singular vectors}. 
We use $k$ to index the incremental matrix, i.e., $\Delta W_k = P_k \Lambda_k Q_k $  for $k = 1, . . . , n$, where n is the number of LoRA layers. We denote the $i$-th singular value of $\Delta W_k$ as $ \lambda_{k,i}$ and the rank of low-rank matrices as $r$. We further denote the parameter sets as $\mathcal{P} = \{P_k\}^n_{k=1}$, $\mathcal{E} = \{\Lambda_k\}^n_{k=1}$, $\mathcal{Q} = \{Q_k\}^n_{k=1}$, and $\mathcal{V} = \{ \mathcal{P}, \mathcal{Q}\}$. To encourage $P_k$ and $Q_k$ to be approximately orthogonal, we use the following regularizer as in  AdaLoRA \citep{zhang2023adaptive}: 
\begin{equation}
R_1=   \sum_{k=1}^{n}(\|P_k^{T}P_k - I\|^{2}_{F} + \|Q_kQ_k^{T} - I\|^{2}_{F}), \label{r1}
\end{equation}
where $I$ is an identity matrix and $\|\cdot\|_{F}$ denotes the Frobenius norm.

\paragraph{Parameterization of Pseudo Singular Values.}
We  parameterize the pseudo singular values in   $\Lambda$ in three specific forms.

\begin{itemize}
    \item \textbf{Real-Value}: All pseudo singular values are real-valued without any constraints. 
    \item \textbf{Softmax}: 
    Given a real vector $v$, we apply the softmax operation to it. $softmax(v)$ are used as the pseudo singular values. 
    These  values 
    add up to one and  represent the contributions of their corresponding singular vector pairs.
    \item \textbf{Approximately Binary}: Given a real vector $v$, we apply element-wise sigmoid to it to transform the values in $v$ into $(0,1)$. Then we use an  element-wise entropy regularizer to encourage the values in  $sigmoid(v)$ are close to either zero or one. The regularizer is defined as:  
\begin{equation}\label{R2}
R_2(\mathcal{E}) = \sum_{k=1}^{n} \sum_{i=1}^{r} \lambda_{k,i} \log\lambda_{k,i} + (1-\lambda_{k,i}) \log(1-\lambda_{k,i}). 
\end{equation}
This setting automatically assigns either a high or low importance to each singular vector pair with the corresponding singular value as zero or one.  
\end{itemize}

\subsection{A Bi-level Optimization Framework}

Our method is based on bi-level optimization, where pseudo singular vector matrices $\mathcal{V}$ and their corresponding pseudo singular value matrices $\mathcal{E}$ are  set as trainable parameters for the lower and upper  level respectively.  

\textbf{Lower Level.} In the lower level, we perform LoRA fine-tuning of a pre-trained model  by minimizing a loss $C$ defined on the first  dataset $D_1$ and low-rank incremental matrices $\{\Delta W_k\}_{k=1}^n$. Calculating $C$ involves  the forward pass for each input example $x$: $W_0x+\Delta Wx=W_0x +P\Lambda Qx$, where $W_0$ is a weight matrix in the pre-trained model. 
$R_1$ in Eq.(\ref{r1}) is applied to promote the approximate orthogonality of $P$ and $Q$. 
The overall training objective is $L_1 = C (\mathcal{V}, \mathcal{E}; D_1) + \gamma_1 R_1(\mathcal{V})$, where $\gamma_1$ is a tradeoff parameter. In this  level, we only train $\mathcal{V}$, while keeping $\mathcal{E}$ tentatively fixed. $\mathcal{E}$ will be updated in the upper level. 
In the end, the  lower level amounts to solving the following problem:
\begin{equation}
\mathcal{V}^{*}(\mathcal{E}) = \mathop{\arg\min}\limits_{\mathcal{V}} \; C (\mathcal{V}, \mathcal{E}; D_1) + \gamma_1 R_1(\mathcal{V}). 
\end{equation}
$\mathcal{V}^{*}(\mathcal{E})$ denotes that the optimal solution $\mathcal{V}^{*}$ depends on $\mathcal{E}$ since $\mathcal{V}^{*}$ depends on $C$ which depends on $\mathcal{E}$.


 

\textbf{Upper Level.} In the upper level, we validate the fine-tuned model where the incremental matrices are parameterized by the  optimally learned  $\mathcal{V}^{*}(\mathcal{E})$ and unlearned pseudo singular values in  $\mathcal{E}$, on the second dataset $D_2$. This results in  a validation  loss $C(\mathcal{V}^{*}(\mathcal{E}),\mathcal{E};D_2)$, which is a function of $\mathcal{E}$. We learn $\mathcal{E}$ by minimizing this loss. Optionally, we use the regularizer $R_2$ in Eq.(\ref{R2}) to encourage the pseudo singular values in $\mathcal{E}$ to be approximately binary. The overall objective function is  $L_2 = C (\mathcal{V}^{*}(\mathcal{E}), \mathcal{E}; D_2 ) + \gamma_2 R_2(\mathcal{E})$, where $\gamma_2$ is a tradeoff parameter. 
This level amounts to solving the following optimization problem: 
\begin{equation}
\mathop{\min}_{\mathcal{E}} \; C(\mathcal{V}^{*}(\mathcal{E}), \mathcal{E}; D_2 ) + \gamma_2 R_2(\mathcal{E}). 
\end{equation}



\begin{algorithm} [t] 
        \caption{BiLoRA} 
        \label{algo}
        \, 1: \textbf{Input}: Datasets $D_1$, $D_2$; unroll steps $T_1$, $T_2$; learning 
        
        \quad\; rates $\eta_1$, $\eta_2$. 

        \, 2: \textbf{In a Global Step} \textbf{do}

        \, 3: \qquad \textbf{for} $t = 1, 2, 3,..., T_1$ \textbf{do}

        \, 4: \qquad \qquad Sample a  minibatch $B_1^{(t)}$ from  $D_1$ 
        
        \, 5: \qquad \qquad Update $\mathcal{V}^{(t)}$ using Eq.(\ref{eq:t1})

        \, 6: \qquad \textbf{for} $t = 1, 2, 3,..., T_2$ \textbf{do}

        \, 7: \qquad \qquad Sample a  minibatch $B_2^{(t)}$ from  $D_2$ 
        
        \, 8: \qquad \qquad  Update $\mathcal{E}^{(t)}$ using using Eq.(\ref{eq:t2})
         
        \, 9: \textbf{end this step}
\end{algorithm}

\textbf{A Bi-level Optimization Framework.} Integrating these two interdependent levels of optimization problems, we have the following bi-level optimization framework: 
$$  \mathop{\min}_{\mathcal{E}} \; C (\mathcal{V}^{*}(\mathcal{E}), \mathcal{E}; D_2 ) + \gamma_2 R_2(\mathcal{E}) $$
$$   s.t. \;  \mathcal{V}^{*}(\mathcal{E}) = \mathop{\arg\min}\limits_{\mathcal{V}} \; C (\mathcal{V}, \mathcal{E}; D_1) + \gamma_1 R_1(\mathcal{V})$$

Note that these two levels of optimization problems are mutually dependent on each other. The output of the lower level, which is $\mathcal{V}^{*}(\mathcal{E})$, is the input of the upper level. The optimization variable $\mathcal{E}$ in the upper level is the input of the lower level. By solving these two interconnected problems jointly, we can learn the pseudo singular vectors and values end-to-end.

\setlength{\tabcolsep}{3pt}
\begin{table*}[t]
\centering
\caption{RoBERTa\textsubscript{base/large} (R\textsubscript{b/l}) with different fine-tuning methods on the GLUE benchmark. We report the average result of five runs with different random seeds.  Higher is better for all metrics. Results of baselines are taken from their original papers. $*$ indicates model already adapted to MNLI when adapting to MRPC, RTE, and STS-B, while $\dagger$ indicates model started as pre-trained when adapting to all datasets.}
\vspace{\baselineskip}

\begin{tabular}{l|r|ccccccccc}
\toprule
Method                                          & Params   & MNLI   & SST-2    & MRPC         & CoLA     & QNLI     & QQP     & RTE    & STS-B  & Avg.     \\ \midrule
R\textsubscript{b}(FT)\textsuperscript{}                           & 125.0M   & 87.6   & 94.8     & 90.2         & 63.6     & 92.8     & \textbf{91.9}    & 78.7   & 91.2   & 86.4     \\ 
R\textsubscript{b}(BitFit)\textsuperscript{}                       &   0.1M   & 84.7   & 93.7     & \textbf{92.7}         & 62.0     & 91.8     & 84.0    & 81.5   & 90.8   & 85.2     \\ 
R\textsubscript{b}(Adpt\textsuperscript{D})\textsuperscript{}      &   0.3M   & 87.1\textsubscript{$\pm$.0}   & 94.2\textsubscript{$\pm$.1}     & 88.5\textsubscript{$\pm$1.1}         & 60.8\textsubscript{$\pm$.4}     & 93.1\textsubscript{$\pm$.1}     & 90.2\textsubscript{$\pm$.0}    & 71.5\textsubscript{$\pm$2.7}   & 89.7\textsubscript{$\pm$.3}   & 84.4     \\ 
R\textsubscript{b}(Adpt\textsuperscript{D})\textsuperscript{}      &   0.9M   & 87.3\textsubscript{$\pm$.1}   & 94.7\textsubscript{$\pm$.3}     & 88.4\textsubscript{$\pm$.1}         & 62.6\textsubscript{$\pm$.9}     & 93.0\textsubscript{$\pm$.2}     & 90.6\textsubscript{$\pm$.0}    & 75.9\textsubscript{$\pm$2.2}   & 90.3\textsubscript{$\pm$.1}   & 85.4     \\ 
R\textsubscript{b}(LoRA)\textsuperscript{$*$}                         &   0.3M   & 87.5\textsubscript{$\pm$.3}   & \textbf{95.1\textsubscript{$\pm$.2}}     & 89.7\textsubscript{$\pm$.7}         & 63.4\textsubscript{$\pm$1.2}     & \textbf{93.3\textsubscript{$\pm$.3}}     & 90.8\textsubscript{$\pm$.1}    & 86.6\textsubscript{$\pm$.7}   & 91.5\textsubscript{$\pm$.2}   & 87.2     \\ 
R\textsubscript{b}(BiLoRA)\textsuperscript{$*$}                         &   0.3M   & \textbf{87.9\textsubscript{$\pm$.2}}   & \textbf{95.1\textsubscript{$\pm$.2}}     & 91.7\textsubscript{$\pm$.5}        & \textbf{64.8\textsubscript{$\pm$.6}}     & \textbf{93.3\textsubscript{$\pm$.2}}     & 91.4\textsubscript{$\pm$.2}    & \textbf{87.2\textsubscript{$\pm$.4}}   & \textbf{91.7\textsubscript{$\pm$.2}}   & \textbf{87.9}     \\ \midrule
R\textsubscript{l}(FT)\textsuperscript{$*$}                        &   355.0M   & 90.2   & 96.4    & 90.9         & 68.0    & 94.7     & \textbf{92.2}    & 86.6   & 92.4   & 88.9     \\ 
R\textsubscript{l}(LoRA)\textsuperscript{$*$}                         &   0.8M   & \textbf{90.6\textsubscript{$\pm$.2}}   & 96.2\textsubscript{$\pm$.5}     & 90.9\textsubscript{$\pm$1.2}         & 68.2\textsubscript{$\pm$1.9}     & 94.9\textsubscript{$\pm$.3}     & 91.6\textsubscript{$\pm$.1}    & 87.4\textsubscript{$\pm$2.5}   & \textbf{92.6\textsubscript{$\pm$.2}}   & 89.0     \\ 
R\textsubscript{l}(BiLoRA)\textsuperscript{$*$}                         &   0.8M   & \textbf{90.6\textsubscript{$\pm$.3}}   & \textbf{96.7\textsubscript{$\pm$.4}}     & \textbf{92.6\textsubscript{$\pm$1.4}}         & \textbf{69.2\textsubscript{$\pm$1.6}}     & \textbf{95.0\textsubscript{$\pm$.1}}     & 92.0\textsubscript{$\pm$.1}    & \textbf{89.5\textsubscript{$\pm$1.1}}   & \textbf{92.6\textsubscript{$\pm$.8}}   & \textbf{89.8} 
\\ \midrule
R\textsubscript{l}(Adpt\textsuperscript{P})\textsuperscript{$\dagger$}      &   3.0M   & 90.2\textsubscript{$\pm$.3}   & 96.1\textsubscript{$\pm$.3}     & 90.2\textsubscript{$\pm$.7}         & 68.3\textsubscript{$\pm$1.0}     & 94.8\textsubscript{$\pm$.2}     & 91.9\textsubscript{$\pm$.1}    & 83.8\textsubscript{$\pm$2.9}   & 92.1\textsubscript{$\pm$.7}   & 88.4    \\ 
R\textsubscript{l}(Adpt\textsuperscript{P})\textsuperscript{$\dagger$}     &   0.8M   & 90.5\textsubscript{$\pm$.3}   & 96.6\textsubscript{$\pm$.2}     & 89.7\textsubscript{$\pm$1.2}         & 67.8\textsubscript{$\pm$2.5}     & 94.8\textsubscript{$\pm$.3}     & 91.7\textsubscript{$\pm$.2}    & 80.1\textsubscript{$\pm$2.9}   & 91.9\textsubscript{$\pm$.4}   & 87.9     \\ 
R\textsubscript{l}(Adpt\textsuperscript{H})\textsuperscript{$\dagger$}      &   6.0M   & 89.9\textsubscript{$\pm$.5}   & 96.2\textsubscript{$\pm$.3}     & 88.7\textsubscript{$\pm$2.9}         & 66.5\textsubscript{$\pm$4.4}     & 94.7\textsubscript{$\pm$.2}     & \textbf{92.1\textsubscript{$\pm$.1}}    & 83.4\textsubscript{$\pm$1.1}   & 91.0\textsubscript{$\pm$1.7}   & 87.8     \\ 
R\textsubscript{l}(Adpt\textsuperscript{H})\textsuperscript{$\dagger$}      &   0.8M   & 90.3\textsubscript{$\pm$.3}   & 96.3\textsubscript{$\pm$.5}     & 87.7\textsubscript{$\pm$1.7}         & 66.3\textsubscript{$\pm$2.0}     & 94.7\textsubscript{$\pm$.2}     & 91.5\textsubscript{$\pm$.1}    & 72.9\textsubscript{$\pm$2.9}   & 91.5\textsubscript{$\pm$.5}   & 86.4     \\ 
R\textsubscript{l}(LoRA)\textsuperscript{$\dagger$}                         &   0.8M   & \textbf{90.6\textsubscript{$\pm$.2}}   & 96.2\textsubscript{$\pm$.5}     & 90.2\textsubscript{$\pm$1.0}         & 68.2\textsubscript{$\pm$1.9}     & 94.8\textsubscript{$\pm$.3}     & 91.6\textsubscript{$\pm$.2}    & 85.2\textsubscript{$\pm$1.1}   & 92.3\textsubscript{$\pm$.5}   & 88.6     \\ 
R\textsubscript{l}(BiLoRA)\textsuperscript{$\dagger$}                         &   0.8M   & \textbf{90.6\textsubscript{$\pm$.3}}   & \textbf{96.7\textsubscript{$\pm$.4}}     & \textbf{92.2\textsubscript{$\pm$1.0}}         & \textbf{69.2\textsubscript{$\pm$1.6}}     & \textbf{95.0\textsubscript{$\pm$.1}}     & 92.0\textsubscript{$\pm$.1}    & \textbf{87.4\textsubscript{$\pm$1.0}}   & \textbf{92.6\textsubscript{$\pm$.8}}   & \textbf{89.5} 
\\ \bottomrule
\end{tabular}
\label{table1}
\vspace{-0.3cm}
\end{table*}

\setlength{\tabcolsep}{3pt}
\begin{table*}[t]
\centering
\caption{DeBERTa-v3-base (D\textsubscript{v3}) with different fine-tuning methods, on the GLUE benchmark. We report the average result of five runs with different random seeds. Higher is better. * indicates results published in prior works.
BiLoRA outperforms FT, LoRA,  AdaLoRA, and other fine-tuning  methods with equal or less parameters.  }

\vspace{\baselineskip}
\begin{tabular}{l|r|ccccccccc}
\toprule
 Method                       & Params                      & MNLI   & SST-2    & MRPC         & CoLA     & QNLI     & QQP     & RTE    & STS-B  & Avg.     \\ \midrule
D\textsubscript{v3}(FT)\textsuperscript{*}                     &184.0M       & 90.01   & 95.63     & 89.46        & 69.19     & 94.03     & 92.40  & 83.75   & 91.60   & 88.09     \\ 
D\textsubscript{v3}(Adpt\textsuperscript{H})\textsuperscript{*}                   &0.6M         & 90.18   & 95.30     & 89.22        & 67.87     & 93.76     & 91.65   & 85.56   & 91.30   & 87.93     \\ 
D\textsubscript{v3}(Adpt\textsuperscript{P})\textsuperscript{*}                   &0.6M         & 90.22   & 95.53     & 89.22        & 69.48     & 93.98     & 91.62   & 84.12   & 91.52   & 88.04     \\ 
D\textsubscript{v3}(LoRA)\textsuperscript{*}                   &0.3M         & 90.34   & 94.95     & 89.71        & 68.71     & 94.03     & 91.61   & 85.56   & 91.68   & 88.15     \\ 
D\textsubscript{v3}(AdaLoRA)\textsuperscript{*}                   &0.3M         & 90.68   & 95.80     & 90.44         & 70.04    & \textbf{94.49}    & 91.78   & 87.36   & 91.63   & 88.86    \\ 
D\textsubscript{v3}(BiLoRA)                    &0.3M       & \textbf{90.81}   & \textbf{96.02}     & \textbf{91.42}         & \textbf{70.52}     & 94.25     & \textbf{91.82}    & \textbf{88.45}   & \textbf{91.96}   & \textbf{89.41}   \\ \bottomrule
\end{tabular}
\label{table2}
\vspace{-0.4cm}
\end{table*}

\paragraph{Optimization Algorithm.} We utilize a gradient-based optimization algorithm~\citep{choe2022betty} to solve this bi-level optimization problem. Our overall optimization algorithm is summarized in Algorithm~\ref{algo}. Specifically, in the lower level, we perform gradient descent for a preset number of steps $T_1$ on the pseudo singular vector matrices $\mathcal{V}$ to approximate the optimal solution $\mathcal{V}^{*}(\mathcal{E})$.  With the initial $\mathcal{V}$ as $\mathcal{V}^{(0)}$ and learning rate $\eta_1$, the gradient descent steps can be formulated as: 
\begin{equation}
    \label{eq:t1}\mathcal{V}^{(t)} = \mathcal{V}^{(t-1)} - \eta_1 \frac{d L_1}{d \mathcal{V}^{(t-1)}},   \text{ for } t=1,2,3,...,T_1.
\end{equation}
We plug $\mathcal{V}^{*}(\mathcal{E}) \approx \mathcal{V}^{(T_1)}$ into the overall objective function in the upper level and get an approximate objective $\widehat{L}_2 = C (\mathcal{V}^{(T_1)}, \mathcal{E}; D_2 ) + \gamma_2 R_2(\mathcal{E})$. We perform gradient descent for a preset number of steps $T_2$ on the pseudo singular values in $\mathcal{E}$ to minimize $\widehat{L}_2$. With the initial $\mathcal{E}$ as $\mathcal{E}^{(0)}$ and learning rate $\eta_2$, the gradient descent steps can be formulated as: 
\begin{equation}
\label{eq:t2}
   \mathcal{E}^{(t)} = \mathcal{E}^{(t-1)} - \eta_2 \frac{d \widehat{L}_2}{d \mathcal{E}^{(t-1)}},   \text{ for } t=1,2,3,...,T_2. 
\end{equation} 
These steps  constitute one global optimization step. We take iterative global steps between the lower level and upper level to solve this bi-level optimization problem until converge.  Following the chain rule, the  hypergradient for the upper  level can be calculated as:
\begin{align*}
 \frac{d \widehat{L}_2}{d \mathcal{E}} = \frac{\partial \widehat{L}_2}{\partial \mathcal{E}} + \frac{\partial \mathcal{V}^{(T_1)}}{\partial \mathcal{E}} \times \frac{\partial \widehat{L}_2}{\partial \mathcal{V}^{(T_1)}}.   
\end{align*}

\section{Experiments}

\label{others}

We evaluated the downstream performance of BiLoRA on RoBERTa \citep{liu2019roberta}, DeBERTa \citep{he2020deberta} and GPT-2 \citep{radford2019language}, and  compared with  LoRA \citep{hu2021lora}, AdaLoRA \citep{zhang2023adaptive},  and other baselines. Our experiments covered a wide range of tasks, from natural language understanding (NLU) to generation (NLG). Specifically, we evaluated RoBERTa and DeBERTa on the GLUE benchmark \citep{wang2018glue} and GPT-2 on the E2E NLG challenge \citep{novikova2017e2e}. We used DeBERTa-xxlarge(1.5B) to evaluate the scaling-up performance of our method. We used NVIDIA A100 for all experiments.

\begin{table*}[t]
\centering
\caption{GPT-2 medium (M) and large (L) with different fine-tuning methods on the E2E NLG Challenge. For all metrics, higher is better. * indicates numbers published in prior works. We keep the same experimental settings as  baselines for a fair comparison.}
\vspace{0.2cm}
\begin{tabular}{l|r|ccccc}
\toprule
Model$\&$Method                                 & Params   & BLEU   & NIST   & MET  & ROUGE-L  & CIDEr         \\ \midrule
GPT-2 M(FT)\textsuperscript{*}                                      & 354.92M  & 68.2   & 8.62   & 46.2 & 71.0 
& 2.47         \\ 
GPT-2 M(Adpt\textsuperscript{L})\textsuperscript{*}                 &   0.37M  & 66.3   & 8.41   & 45.0 & 69.8     & 2.40         \\ 
GPT-2 M(Adpt\textsuperscript{L})\textsuperscript{*}                 &  11.09M  & 68.9   & 8.71   & 46.1 & 71.3     & 2.47         \\ 
GPT-2 M(Adpt\textsuperscript{H})\textsuperscript{*}                 &  11.09M  & 67.3\textsubscript{$\pm$.6}   & 8.50\textsubscript{$\pm$.07}   & 46.0\textsubscript{$\pm$.2} & 70.7\textsubscript{$\pm$.2}    & 2.44\textsubscript{$\pm$.01}        \\ 
GPT-2 M(FT\textsuperscript{Top2})\textsuperscript{*}                &  25.19M  & 68.1   & 8.59   & 46.0 & 70.8     & 2.41          \\ 
GPT-2 M(PreLayer)\textsuperscript{*}                                &   0.35M  & 69.7   & 8.81   & 46.1 & 71.4     & 2.49          \\ 
GPT-2 M(LoRA)\textsuperscript{*}                                    &   0.35M  & 70.4\textsubscript{$\pm$.1}   & 8.85\textsubscript{$\pm$.02}  & 46.8\textsubscript{$\pm$.2} & 71.8\textsubscript{$\pm$.1}    & 2.53\textsubscript{$\pm$.02}         \\ 
GPT-2 M(BiLoRA)                                                       &   0.35M  & \textbf{70.8\textsubscript{$\pm$.4}}   & \textbf{8.87\textsubscript{$\pm$.03}}   & \textbf{46.9\textsubscript{$\pm$.1}} & \textbf{72.1\textsubscript{$\pm$.2}}     & \textbf{2.55\textsubscript{$\pm$.03}}  \\ \midrule 
GPT-2 L(FT)\textsuperscript{*}                                      & 774.03M  & 68.5   & 8.78   & 46.0 & 69.9     & 2.45         \\ 
GPT-2 L(Adpt\textsuperscript{L})\textsuperscript{*}                 &   0.88M  & 69.1\textsubscript{$\pm$.1}   & 8.68\textsubscript{$\pm$.03}  & 46.3\textsubscript{$\pm$.0}& 71.4\textsubscript{$\pm$.2}     & \textbf{2.49\textsubscript{$\pm$.0}}         \\ 
GPT-2 L(Adpt\textsuperscript{L})\textsuperscript{*}                 &  23.00M  & 68.9\textsubscript{$\pm$.3}   & 8.70\textsubscript{$\pm$.04}   & 46.1\textsubscript{$\pm$.1} & 71.3\textsubscript{$\pm$.2}     & 2.45\textsubscript{$\pm$.02}          \\ 
GPT-2 L(PreLayer)\textsuperscript{*}                                &   0.77M  & 70.3   & 8.85   & 46.2 & 71.7     & 2.47          \\ 
GPT-2 L(LoRA)\textsuperscript{*}                                    &   0.77M  & 70.4\textsubscript{$\pm$.1}   & 8.89\textsubscript{$\pm$.02}   & 46.8\textsubscript{$\pm$.2}&  \textbf{72.0\textsubscript{$\pm$.2}}     & 2.47\textsubscript{$\pm$.02}          \\ 
GPT-2 L(BiLoRA)                                                       &   0.77M  & \textbf{70.6\textsubscript{$\pm$.3}}   & \textbf{8.91\textsubscript{$\pm$.04}}   & \textbf{47.0\textsubscript{$\pm$.3}} &\textbf{ 72.0\textsubscript{$\pm$.4}}     & \textbf{2.51\textsubscript{$\pm$.03}} \\  \bottomrule
\end{tabular}
\label{table3}
\vspace{-0.4cm}
\end{table*}

\subsection{Baselines}

We compared with the same baselines as LoRA and AdaLoRA, and used the reported results in previous works. Additionally, we also took LoRA and AdaLoRA as our baselines to evaluate the effectiveness of our method.

\textbf{Full Fine-Tuning (FT)} is a frequently employed method for adaptation. The model is initialized with pre-trained weights and biases and all model parameters are subjected to gradient updates. We also included a simple variant reported in prior work on GPT-2 \citep{li2021prefix}, which only adapts the last two layers while freezing others.

\textbf{Bias-only or BitFit} \citep{zaken2021bitfit} is an effective PEFT method which only trains the bias vectors while freezing everything else in the pre-trained model.

\textbf{Prefix-embedding tuning (PreEmbed)} introduces specialized tokens within the input tokens, featuring trainable word embeddings that typically do not belong to the model's vocabulary \citep{li2021prefix}. 

\textbf{Prefix-layer tuning (PreLayer)} learns the activations after every Transformer layer by replacing the activations computed from previous layers with trainable parameters. This method can be seen as an extension to prefix-embedding tuning.

\textbf{Adapter tuning} \citep{houlsby2019parameter} inserts layer-adapters between neural modules such as the MLP module or the self-attention module. We used  four types of adapters as in LoRA \citep{hu2021lora}:  \textbf{Adapter\textsuperscript{L}} with the adapter layer applied only after the MLP module and after a LayerNorm \citep{lin2020exploring}, \textbf{Adapter\textsuperscript{D}} with some adapter layers dropped for increasing efficiency \citep{ruckle2020adapterdrop}. \textbf{Adapter\textsuperscript{H}} incorporates two fully connected layers within an adapter layer, with  nonlinearity in between \citep{houlsby2019parameter}. \textbf{Adapter\textsuperscript{P}} \citep{pfeiffer2020adapterfusion} is similar to \textbf{Adapter\textsuperscript{L}},  but introduces a novel two-stage transfer learning strategy to combine the knowledge from multiple source tasks.

\textbf{LoRA} \citep{hu2021lora} adds trainable incremental update matrices to pre-trained weight matrices. Following the experimental settings of LoRA, we applied BiLoRA to  $W_q$ and $W_v$ matrices (the query and value weight matrices in the self-attention module) for a fair comparison. 


\textbf{AdaLoRA} \citep{zhang2023adaptive} proposes SVD-based adaptation and rank-allocation, which formulates the incremental matrices in the form of singular value decomposition and allocates rank budget based on importance scores.

\subsection{Natural Language Understanding}

For natural language understanding (NLU) tasks, we conducted  experiments on the General Language Understanding Evaluation (GLUE) benchmark for RoBERTa and DeBERTa. Please see   Appendix~\ref{appena} for more details on the models and datasets we use and see Appendix~\ref{appeng} for more comparisons and results.

\textbf{Implementation Details.}
Our implementation is based on \textit{Huggingface Transformers} \citep{wolf2019huggingface} and \textit{Betty} \citep{choe2022betty}. \textit{Betty} is a software library for solving large-scale multilevel optimization (MLO) problems. Specifically, we load RoBERETa and DeBERTa models with \textit{Huggingface Transformers} and build our bi-level optimization framework with \textit{Betty}.

\textbf{Experimental Settings.}
Following LoRA, we used the development set in GLUE as test data since the test set is not publicly available. We divided the training set into 
two datasets, with an 8:2 split, serving as the lower-level and upper-level datasets respectively in our bi-level formulation. We maintained this fixed ratio for all tasks. 
Singular values were  parameterized as Softmax if not otherwise stated  and $R_1$ was added to the lower level as a regularizer. 
For RoBERTa base/large, we kept our experimental settings the same as LoRA. 
For DeBERTa-v3-base, we kept our experimental settings close to AdaLoRA while maintaining a lower parameter budget. 
We also kept  hyperparameters such as sequence length, total batch size, LoRA rank,  and LoRA alpha exactly the same as LoRA/AdaLoRA where necessary. These experimental settings allow for a fair comparison with all baseline methods. Please see the Appendix \ref{appenb} for all the hyperparameter settings. The role and choice of unroll step $T_1, T_2$ and data partition $D_1, D_2$ are further analyzed in Appendix~\ref{appenf}.

\begin{table}[t] \centering
\caption{GPT-2 medium (M) with different fine-tuning methods on the WebNLG and DART datasets. We reported the BLEU score and higher is better. * indicates numbers published in prior works.}
\vspace{0.2cm}
\begin{tabular}{l|ccccccccc}
\toprule
 Method   &  WebNLG	      & DART    \\ \midrule
GPT-2 M(FT)\textsuperscript{*}  	&46.5	&46.2 \\
GPT-2 M(Adpt\textsuperscript{L})\textsuperscript{*} 	&50.2	&42.4 \\
GPT-2 M(FT\textsuperscript{Top2})\textsuperscript{*}	&54.9	&41.0\\
GPT-2 M(PreLayer)\textsuperscript{*}  	&36.0	&46.4\\
GPT-2 M(LoRA)\textsuperscript{*}  	&55.3	&47.1\\
GPT-2 M(AdaLoRA)  	&55.5	&47.6\\
GPT-2 M(BiLoRA) 	&\textbf{56.1}	&\textbf{49.0} \\
 \bottomrule
\end{tabular}
\label{table4}
\vspace{-0.5cm}
\end{table}

\begin{table}[t] \centering
\caption{Experiment results for scaling up to DeBERTa-XXL (D\textsubscript{v2}). In BiLoRA, the   values of hyperparameters including  LoRA rank, LoRA alpha, and max length are the same as those in LoRA. * indicates numbers published in prior works.  }
\vspace{\baselineskip}
\begin{tabular}{l|r|cccccc}
\toprule
 Method                                        &params     & MNLI       & MRPC         & CoLA           & Avg.     \\ \midrule
D\textsubscript{v2}(FT)\textsuperscript{*}                     &1500.0M       & 91.8   & 92.0         & 72.0        & 85.3     \\ 
D\textsubscript{v2}(LoRA)\textsuperscript{*}                   &4.7M         & \textbf{91.9\textsubscript{$\pm$.2}}   & 92.6\textsubscript{$\pm$.6}         & 72.4\textsubscript{$\pm$1.1}        & 85.6     \\ 
D\textsubscript{v2}(BiLoRA)                    &4.7M       & \textbf{91.9\textsubscript{$\pm$.3}}   & \textbf{92.7\textsubscript{$\pm$.4}}     & \textbf{73.0\textsubscript{$\pm$.4}}     & \textbf{85.9}        \\ \bottomrule
\end{tabular}
\label{table5}
\end{table}

\textbf{Main Results.}
The same as LoRA, we report the overall (matched and mismatched) accuracy for MNLI, Matthew's correlation for CoLA, Pearson correlation for STS-B, and accuracy for the other tasks. 
Table~\ref{table1} shows the results of RoBERTa base/large on the GLUE development set. As can be seen, our method outperforms LoRA on all datasets with the same number of trainable parameters. On most datasets, our method achieves better or on par performance compared with  baselines. The average score of BiLoRA notably outperforms all the baselines. Table~\ref{table2} shows the results of DeBERTa-v3-base on the GLUE development set. BiLoRA outperforms all baselines with equal or less trainable parameters. The improvements achieved by our method over baselines are attributed to its bi-level learning  mechanism which separates the training of pseudo singular vectors and values on two distinct sub-datasets. As a result, it  effectively alleviates the risk of overfitting to one dataset and yields better generalization performance. In contrast, baseline methods train all parameters on the same dataset and thus lead to overfitting to this dataset. 
This is particularly evidenced by the observation  that 
on smaller datasets such as CoLA, RTE, and MRPC where overfitting is more likely to occur, BiLoRA 
outperforms  baselines by a larger margin.

\subsection{Natural Language Generation}

For natural language generation (NLG) tasks, we followed the setup of Prefix-Tuning \citep{li2021prefix} and LoRA \citep{hu2021lora} on GPT-2 for a direct comparison with LoRA and other fine-tuning  methods. We evaluated GPT-2 medium and large on the E2E NLG Challenge. Please see  Appendix~\ref{appena} for more details on the models and datasets we used and see Appendix~\ref{appeng} for more comparisons and results.

\textbf{Implementation Details.} 
Our implementation is based on the fine-tuning code for GPT-2 in Huggingface and Betty  \citep{choe2022betty}. Specifically, we load GPT-2 models with the code of Huggingface and build our bi-level optimization framework with Betty.

\begin{table}[t] \centering
\caption{Experiment results on three different parameterizations of pseudo singular values: Real Value, Softmax, and Approximately Binary. 
}
\vspace{0.2cm}
\resizebox{0.48\textwidth}{!}{
\begin{tabular}{l|ccccccccc}
\toprule
 Method                                             & MNLI   & SST-2    & MRPC         & CoLA     & QNLI     & QQP     & RTE    & STS-B  & Avg.     \\ \midrule
 R\textsubscript{b}(LoRA)                           & 87.5   & \textbf{95.1}     & 89.7         & 63.4     & \textbf{93.3}     & 90.8    & 86.6   & 91.5   & 87.2     \\ 
 R\textsubscript{b}(Real Value)                           & 87.5   & 94.6     & 91.7         & 63.6     & 93.0     & 90.8    & 86.6   & 91.3   & 87.4     \\ 
R\textsubscript{b}(Softmax)                           & \textbf{87.9} & \textbf{95.1}     & \textbf{91.7}         & \textbf{64.8}     & \textbf{93.3}     & \textbf{91.4}    & \textbf{87.2}   & \textbf{91.7}   & \textbf{87.9}     \\ 
R\textsubscript{b}(Binary)                          & 87.6   & 94.8     & 91.4         & 64.4     & 93.0     & 91.2    & 86.6   & 91.5   & 87.6     \\ 

 \bottomrule
\end{tabular}
}
\label{table6}
\vspace{-0.3cm}
\end{table}

\textbf{Experimental Settings.}
In our method, the  training set and validation set are used  as the lower-level and upper-level datasets respectively, and we report performance on the test set. Singular values were  parameterized as Softmax if not otherwise stated.
We kept our experimental settings the same as LoRA. Specifically, we kept hyperparameters such as sequence length, batch size, LoRA rank, LoRA  alpha, and label smoothing  exactly the same as LoRA. These experimental settings allow for a fair comparison with LoRA and other fine-tuning  methods.

\textbf{Main Results.} 
Table~\ref{table3} and Table~\ref{table4} show the results of GPT-2 medium/large on the E2E test set and GPT-2 medium on WebNLG and DART test sets.  Our method outperforms LoRA and other methods on all metrics  for both GPT-2 M and GPT-2 L. The results  demonstrate the effectiveness of our method in Natural Language Generation (NLG) downstream  tasks and the generalization capabilities of our method across different models and task types. 


\begin{table}[t] \centering
\caption{Experiment results of RoBERTa\textsubscript{base} (R\textsubscript{b}) on GLUE, under different values of $\gamma_1$. 
}
\vspace{0.2cm}
\resizebox{0.48\textwidth}{!}{
\begin{tabular}{l|ccccccccc}
\toprule
 Method                                             & MNLI   & SST-2    & MRPC         & CoLA     & QNLI     & QQP     & RTE    & STS-B  & Avg.     \\ \midrule
R\textsubscript{b}($\gamma_1=0.0$)                           & 87.8   & 95.0    & 91.7         & \textbf{64.8}    & 93.1    & \textbf{91.5}    & 87.2   & \textbf{91.7}   & \textbf{87.9}     \\ 
R\textsubscript{b}($\gamma_1=0.1$)                           & \textbf{87.9}   & \textbf{95.1}     & 91.7         & \textbf{64.8}    & \textbf{93.3}     & 91.4    & 87.2   & \textbf{91.7}   & \textbf{87.9}     \\ 
R\textsubscript{b}($\gamma_1=0.2$)                          & 87.8   & 95.0     & \textbf{91.9}         & 64.4     & 93.1     & 91.2    & 86.9   & 91.5   & 87.7    \\ 
R\textsubscript{b}($\gamma_1=0.3$)                           & 87.2   & 94.6    & 91.4        & 63.6     & 92.8     & 90.9    &\textbf{87.4}   & 91.2   & 87.4     \\ \bottomrule
\end{tabular}
}
\label{table7}
\vspace{-0.2cm}
\end{table}

\subsection{Analysis}

\paragraph{Scaling Up to DeBERTa-XXL.} 
We use DeBERTa-v2-xxlarge(1.5B) to evaluate the scaling-up performance of our method. The study was performed on three datasets of the GLUE benchmark due to the constraint of computational resources for keeping the same experimental settings as LoRA. Results in  Table~\ref{table5} show that BiLoRA achieves better or on par performance compared with LoRA and full fine-tuning (FT), indicating that BiLoRA yields better generalization  when applied for fine-tuning models with a very large number of parameters. 

\paragraph{Ablation Studies on Pseudo Singular Values.}
In Section~\ref{sec:para}, we introduced three ways to parameterize the pseudo singular values: Real Value, Softmax, and Approximately Binary. We conduct experiments separately using these three parameterization methods while keeping other experimental settings the same. We test RoBERTa's performance on the GLUE dataset. Results  in Table~\ref{table6} show that the Softmax parameterization exhibits the best performance, with Approximately Binary coming in a close second. Softmax and Approximately Binary outperform Real Value  because they yield  positive values which meet  the constraint that singular values need to be non-negative while Real Value does not.  
Approximately Binary performs slightly worse than  Softmax since it imposes a stronger constraint that the values need to be close to zero or one. Such a constraint limits the expressivity of the  parameterization. Another observation is that under all the three parameterization methods, BiLoRA outperforms LoRA,  demonstrating  that BiLoRA is robust against different ways of representing the pseudo singular values and thus does not require extensive tuning for selecting the best parameterization. We further provided the  distribution and analysis of singular values for BiLoRA and AdaLoRA in Appendix~\ref{append} to offer additional  insights.

\paragraph{Ablation Study on Orthogonality-Promoting  Regularization.}
We investigated how the tradeoff parameter $\gamma_1$ associated with the orthogonality-promoting regularizer $R_1$ in Eq.(\ref{r1}) affects the performance of our method. The study was performed on RoBERTa-base.  
Results in Table~\ref{table7} show that our method is robust against different values of $\gamma_1$, which implies that using our method does  not need to extensively tune this hyperparameter. We further illustrated the orthogonality of singular vectors and analyzed the reason for the robustness in Appendix~\ref{appene}.

\paragraph{Computation Costs.} 
Table~\ref{table8} shows the training time of LoRA  and our method. The total training time of our method on the eight datasets is lower than that of LoRA. 
 This arises from the fact that BiLoRA converges with much fewer training epochs than LoRA. 
In the Softmax parameterization of pseudo singular values, each value is initialized with a mean equal to $1/r$, larger than that in Real-Value, which increases the overall magnitude of $\Delta W$ and allows a larger learning rate for the training process. 
The bi-level optimization framework effectively accommodates this larger  learning rate by iteratively optimizing between the two levels without affecting the training stability. With such a large learning rate, even though bi-level optimization takes longer time for each training step, it takes much fewer training steps for training low-rank matrices compared to LoRA, thus reducing the total training time. We further compared the total step, total time and per-step cost of LoRA, AdaLoRA and BiLoRA in Appendix~\ref{appeng}.

\paragraph{Other Methods Targeting Overfitting.} There are some common experimental settings often used for mitigating overfitting. For AdaLoRA, two promising methods are increasing weight decay and adopting a more aggressive rank pruning setting. Results in Table~\ref{tableweight} show that the application of an increased weight decay to AdaLoRA results in a decline in the overall performance. We further investigated the effect of rank pruning settings and illustrated the results through loss curves in Appendix~\ref{appenc}.
Both experiment results and loss curves indicate that neither of these two approaches effectively addresses the overfitting issue nor enhances the model’s generalization ability in these experiments, which necessitates BiLoRA as a novel and efficient method for mitigating overfitting in this regard.

\begin{table}[t] \centering

\caption{Training time (minutes) of LoRA and BiLoRA on RoBERTa\textsubscript{base/large} (R\textsubscript{b/l}) and 
the GLUE benchmark.  
 }
 \vspace{0.2cm}
 \resizebox{0.48\textwidth}{!}{
\begin{tabular}{l|ccccccccc}
\toprule
 Method                                             & MNLI   & SST-2    & MRPC         & CoLA     & QNLI     & QQP     & RTE    & STS-B   &Total.    \\ \midrule
R\textsubscript{b}(LoRA)                            & 3190.7   & 1096.2     & \textbf{30.2}           & \textbf{193.0}   &709.8     & 2464.3    & \textbf{55.5}   & \textbf{62.4}     & 7802.1  \\ 
R\textsubscript{b}(BiLoRA)                            & \textbf{1407.1}   & \textbf{260.1}     & 240.3         &260.3     &\textbf{375.2}     & \textbf{1732.6}    & 97.5   & 158.3        & \textbf{4531.4} \\  \midrule
R\textsubscript{l}(LoRA)                           & 789.7   & \textbf{133.9}     & \textbf{14.7}         & \textbf{34.1}     & 209.1     & 1446.7    & 10.0   & \textbf{23.1}    & 2661.3  \\ 
R\textsubscript{l}(BiLoRA)                           & \textbf{707.5}   & 160.8      & 19.2         & 62.5     & \textbf{200.4}    &\textbf{1166.7}     &\textbf{4.4}    & 43.3  &  \textbf{2363.8}    \\ \bottomrule
\end{tabular}
}
\label{table8}
\vspace{-0.2cm}
\end{table}

\begin{table}[t] \centering
\caption{DeBERTa-v3-base (D\textsubscript{v3}) with different weight decays for AdaLoRA on the GLUE benchmark. All other hyperparameters are kept the same. }
\vspace{0.2cm}
\begin{tabular}{c|ccccccccc}
\toprule
 Weight Decay                    &SST-2      & CoLA   &QNLI   \\ \midrule
 0.00        & 95.80          & \textbf{70.04}   & \textbf{94.49}        \\ 
 0.05        & \textbf{95.96}  & 68.03         &  94.03                 \\ 
 0.10       &  95.30         & 68.42          &   93.52               \\ 
 0.20       &  94.60         & 67.87          &   93.76               \\ 
 \bottomrule
\end{tabular}
\label{tableweight}
\vspace{-0.2cm}
\end{table}

The results above jointly  demonstrate that BiLoRA enhances training performance while reducing the overall training time. These results substantiate the effectiveness of our method.

\section{Conclusion and Future Work}
We propose BiLoRA, a novel and general bi-level optimization framework for  enhancing the performance of LoRA methods   through addressing their  overfitting issue. By utilizing the SVD parameterization form of low-rank incremental matrices, our method separately trains pseudo singular vectors and singular values on different sub-datasets in two different optimization levels. Such a method effectively alleviates overfitting  while reducing the total training time, as demonstrated in extensive experiments on NLU and NLG tasks.  


Our method opens up several potential directions for future research: 1) The parameterization form of pseudo singular values can be further developed  to support automated rank selection. 2) Our bi-level optimization framework enhances the generalization capability of fine-tuned models, which encourages further in-depth theoretical analysis in this regard.

\clearpage

\section*{Impact Statements}
This paper proposes a bi-level optimization (BLO) method to combat overfitting in low-rank adaptation (LoRA) during fine-tuning of large-scale pre-trained models. This approach enhances model generalization in natural language tasks, contributing to the advancement of Machine Learning with potential implications for improved model capability across applications.



\bibliography{example_paper}
\bibliographystyle{icml2024}

\newpage
\appendix
\onecolumn
\section{Datasets and Models}\label{appena}

\subsection{Natural Language Understanding}
\textbf{GLUE Benchmark} comprises a diverse array of natural language understanding tasks widely employed for evaluation. It encompasses two single-sentence classification tasks, three tasks assessing similarity and paraphrasing, and four tasks focusing on natural language inference.  Specifically, it includes MNLI (MultiNLI, \citet{williams2017broad}), SST-2 (Stanford Sentiment Treebank, \citet{socher2013recursive}), MRPC (Microsoft Research Paraphrase Corpus, \citet{dolan2005automatically}), CoLA (Corpus of Linguistic Acceptability, \citet{warstadt2019neural}), QNLI (Question NLI, \citet{rajpurkar2018know}), QQP (Quora Question Pairs), RTE (Recognizing Textual Entailment), and STS-B (Semantic Textual Similarity Benchmark, \citet{cer2017semeval}). We summarized the statistical data for all datasets within the GLUE Benchmark in the table below:

\setlength{\tabcolsep}{3pt}
\begin{table}[h] \centering
\caption{The statistical data for all datasets within the GLUE Benchmark}
\vspace{\baselineskip}
\begin{tabular}{l|r|c|c|c|c|c}

\toprule
Dataset          & Metrics          & Train   & Dev      & Test          & Label      &Task         \\ \midrule
MNLI             & Accuracy         & 393k    &20k       &20k            &3           & NLI        \\ \midrule
SST-2            & Accuracy         & 67k     &872       &1.8k           &2           & Sentiment     \\ \midrule
MRPC             & Accuracy         & 3.7k    &408       &1.7k           &2           & Paraphrase        \\ \midrule
CoLA             & Matthews corr    & 8.5k    &1k        &1k             &2           & Acceptability     \\ \midrule
QNLI             & Accuracy         & 108k    &5.7k      &5.7k           &2           & QA/NLI        \\ \midrule
QQP              & Accuracy         & 364k    &40k       &391k           &2           & Paraphrase        \\ \midrule
RTE              & Accuracy         & 2.5k    &276       &3k             &2           & NLI        \\ \midrule
STSB             & Pearson corr     & 7.0k    &1.5k      &1.4k           &1           & Similarity        \\ 
\bottomrule
\end{tabular}
\label{hyperroberta}
\end{table}

\subsection{Natural Language Generation}
\textbf{E2E NLG Challenge} \citep{novikova2017e2e} is now commonly used for data-to-text evaluation. It was first introduced as a dataset for training end-to-end, data-driven natural language generation systems. Multiple references can be associated with each source table used as input. Each sample input $(x, y)$ is composed of a series of slot-value pairs, accompanied by an associated natural language reference text. The E2E dataset consists of approximately 42,000 training examples, 4,600 validation examples, and 4,600 test examples from the restaurant domain.

\subsection{Models}

\textbf{RoBERTa} \citep{liu2019roberta} builds upon the foundational principles and training strategies of BERT \citep{devlin2018bert}, offering novel alternatives that enhance downstream task performance. RoBERTa refines and optimizes the pre-training methodology initially proposed in BERT, resulting in notable improvements in task performance while maintaining a comparable number of trainable parameters. We use RoBERTa-base and RoBERTa-large for a convenient and fair comparison with LoRA \citep{hu2021lora}.

\textbf{DeBERTa} \citep{he2020deberta} represents an advanced iteration of BERT, having undergone extensive training at a larger scale. DeBERTa demonstrates strong competitiveness when evaluated on the GLUE benchmark. For our experiments, we use DeBERTa-v2-xxlarge which has 1.5 billions of parameters to evaluate the scaling-up capability of BiLoRA and also for a convenient comparison with LoRA. We use DeBERTa-v3-base which has 183 millions parameters for fair comparison with AdaLoRA \citep{zhang2023adaptive}.

\textbf{GPT-2} \citep{radford2019language} developed by OpenAI, was once a state-of-the-art language model renowned for its remarkable text generation capabilities. It is a scaled-up version of its predecessor, GPT-1, and is trained on an extensive corpus of text data. GPT-2 has been widely recognized for its proficiency in generating coherent and contextually relevant text across various natural language understanding and generation tasks, showcasing its versatility and potential in the field of natural language processing.

\section{Experimental Settings}\label{appenb}

\subsection{RoBERTa}
We summarized the experimental settings for the experiments of RoBERTa-base and RoBERTa-large in Table\ref{roberthy}. In fact, we only introduced an additional level of learning rate compared to LoRA. For hyperparameters such as max seq length, LoRA $\alpha$, we kept them the same as LoRA. We chose learning rates from the magnitude of 1e-5 for almost all of our experiments.  The hyperparameter tuning for our method is quite simple, convenient and straightforward.

\setlength{\tabcolsep}{3pt}
\begin{table}[h] \centering 
\caption{The hyperparameters we used for RoBERTa on the GLUE benchmark. $*$ indicates model already adapted to MNLI when adapting to MRPC, RTE, and STS-B, while $\dagger$ indicates model started as pre-trained when adapting to all datasets.}
\vspace{\baselineskip}
\begin{tabular}{ll|cccccccc}

\toprule
 Method                                          & Settings   & MNLI   & SST-2    & MRPC         & CoLA     & QNLI     & QQP     & RTE    & STS-B       \\ \midrule
        & Optimizer  & \multicolumn{8}{c}{AdamW}   \\ 
        & Warmup Ratio   & \multicolumn{8}{c}{0.06}         \\ 
        & Scheduler   & \multicolumn{8}{c}{Linear}         \\ 
        & LoRA rank   & \multicolumn{8}{c}{$rank_q = rank_v = 8$}         \\  \midrule
RoBERTa-base\textsuperscript{*}        & Total batch size   & \multicolumn{8}{c}{64}         \\ 
                    & Global steps          & 10k  & 3k   &2k   &3k   &3k   &15k  &1.5k &5k      \\ 
                    & Lower learning rate   & 2e-5 &3e-5  &2e-6 &2e-5  &3e-5 &3e-5 &4e-6 &4e-6 \\ 
                    & Upper learning rate   & 3e-5 &4e-5  &8e-6 &4e-5  &4e-5 &4e-5 &2e-6 &2e-6 \\ 
                    & Lower weight decay    & 0.12 &0.12  &0.12 &0.12  &0.12 &0.12 &0.12 &0.1 \\ 
                    & Upper weight decay    & 0.1  &0.1   &0.1  &0.1 &0.1 &0.1 &0.1 &0.1 \\ 
                    & Max Seq Length  & \multicolumn{8}{c}{512}         \\ \midrule
RoBERTa-large\textsuperscript{*}       & Total batch size   & \multicolumn{8}{c}{32}         \\ 
                    & Global steps          & 15k  & 4k   &1k       &3k     &3k   &20k    &0.12k  &2k      \\ 
                    & Lower learning rate   & 1.5e-5 &1.5e-5  &4e-6   &1e-5   &1e-5    &1e-5   &4e-6  &1e-5\\ 
                    & Upper learning rate   & 2e-5 &2e-5  &6e-6   &5e-5   &3e-5     &2e-5   &4e-6   &5e-6\\ 
                    & Lower weight decay    & 0.12 &0.12  &0.12   &0.12   &0.12    &0.12   &0.1    &0.1\\ 
                    & Upper weight decay    & 0.1  &0.1   &0.1   &0.1    &0.1    &0.1   &0.1      &0.1\\ 
                    & Max Seq Length  & \multicolumn{8}{c}{128}         \\ \midrule
RoBERTa-large\textsuperscript{$\dagger$}       & Total batch size   & \multicolumn{8}{c}{32}         \\ 
                    & Global steps          & 15k  & 4k   &1k       &3k     &3k   &20k    &2k   & 2k     \\ 
                    & Lower learning rate   & 1.5e-5 &1.5e-5  &2e-5   &1e-5   &1e-5    &1e-5   &1e-5   &8e-6 \\ 
                    & Upper learning rate   & 2e-5 &2e-5  &1e-4   &5e-5    &3e-5    &2e-5   &2e-5    &4e-6\\ 
                    & Lower weight decay    & 0.12 &0.12  &0.12   &0.12   &0.12    &0.12   &0.12   &0.1\\ 
                    & Upper weight decay    & 0.1  &0.1   &0.1   &0.1    &0.1      &0.1   &0.1    &0.1\\ 
                    & Max Seq Length  & \multicolumn{8}{c}{128}         \\ \midrule
\end{tabular}
\label{roberthy}
\end{table}

\subsection{DeBERTa}

We summarized the experimental settings used in the experiments for DeBERTa-v2-xxlarge and DeBERTa-v3-base in Table\ref{deberthy}. In fact, we only introduced an additional level of learning rate compared to LoRA. For hyperparameters such as max seq length, LoRA $\alpha$, we kept them the same as LoRA and AdaLoRA. We chose learning rates from the magnitude of 1e-5 for almost all of our experiments. The hyperparameter tuning for our method is quite simple, convenient and straightforward. Due to our limited computational resources, we were unable to maintain the same experimental settings as LoRA on many datasets, making a fair comparison impossible. Therefore, for DoBERTa-v2-xxlarge, we only conducted experiments on the MNLI, CoLA, and MRPC datasets.

\setlength{\tabcolsep}{3pt}
\begin{table}[t] \centering 
\setlength{\abovecaptionskip}{-0.2cm}
\setlength{\belowcaptionskip}{\baselineskip}
\caption{The hyperparameters we used for DeBERTa-v2-xxlarge and DeBERTa-v3-base on the GLUE benchmark.}
\begin{tabular}{ll|cccccccc}

\toprule
 Method                                          & Settings   & MNLI   & SST-2    & MRPC         & CoLA     & QNLI     & QQP     & RTE    & STS-B       \\ \midrule
        & Optimizer  & \multicolumn{8}{c}{AdamW}   \\ 
        & Scheduler   & \multicolumn{8}{c}{Linear}         \\ 
        & LoRA rank   & \multicolumn{8}{c}{$rank_q = rank_v = 8$}         \\  \midrule
DeBERTa-v2-XXL          & Total batch size      & 64     &    &32   &32   &    &   &  &        \\ 
                        & Global steps          & 20k    &    &1k   &3k   &    &   &  &        \\ 
                        & Inner learning rate   & 0.5e-5 &    &2e-6   &1e-5   &    &   &  &       \\ 
                        & Outer learning rate   & 1e-5   &    &2e-6   &1e-5   &    &   &  &        \\ 
                        & LoRA $\alpha$         & 16     &    &16   &16   &    &   &  &     \\
                        & Max Seq Length        & 128    &    &128   &64   &    &   &  &         \\ \midrule
DeBERTa-v3-base     & Total batch size   & \multicolumn{8}{c}{32}         \\ 
                        & Global steps          & 15k  & 3k   &1k   &1k   &2k    &20k   &0.5k  &1k       \\ 
                        & Lower learning rate   & 1e-5 &1.5e-5&2e-6 &1e-5 &1e-5  &1e-5 &4e-6  &4e-6  \\ 
                        & Upper learning rate   & 2e-5 &2.5e-5&4e-6 &2e-4 &2e-5  &2e-5 &4e-6  &4e-6  \\ 
                        & Lower weight decay    & 0.12 &0.12  &0.15 &0.12 &0.12  &0.12 &0.12  &0.12  \\ 
                        & Upper weight decay    & 0.1  &0.1   &0.1  &0.1  &0.1   &0.1  &0.1   &0.1  \\ 
                        & LoRA $\alpha$   & \multicolumn{8}{c}{16}         \\
                        & Max Seq Length  & 256  &128   &320  &64  &512 &320 &320 &128         \\  \bottomrule
\end{tabular}
\label{deberthy}
\end{table}

\subsection{GPT-2}
We summarized the experimental settings for the experiments of GPT-2 M and L in Table\ref{gpthy}. We kept hyperparameters almost the same as LoRA for a fair comparison.

\setlength{\tabcolsep}{3pt}
\begin{table}[h] \centering 
\setlength{\abovecaptionskip}{-0.2cm}
\setlength{\belowcaptionskip}{\baselineskip}
\caption{The hyperparameters we used for GPT-2 on the E2E NLG benchmark. }
\begin{tabular}{l|c}

\toprule
Settings    &  Training    \\ \midrule
 Optimizer  & \multicolumn{1}{c}{AdamW}   \\ 
 Warmup Steps   & \multicolumn{1}{c}{500}         \\ 
 Scheduler   & \multicolumn{1}{c}{Linear}         \\ 
 LoRA rank   & \multicolumn{1}{c}{$rank_q = rank_v = 4$}         \\  
 LoRA $\alpha$ & \multicolumn{1}{c}{32}  \\
 Label Smooth &  \multicolumn{1}{c}{0.1}  \\
 Weight Decay &  \multicolumn{1}{c}{0.01} \\
 Batch Size &  \multicolumn{1}{c}{8}  \\ \midrule
Settings      & Inference   \\ \midrule
Beam Size & \multicolumn{1}{c}{10}   \\
 Length Penalty & \multicolumn{1}{c}{0.9}   \\
 no repeat ngram size & \multicolumn{1}{c}{4}   \\ \bottomrule
\end{tabular}
\label{gpthy}
\end{table}


\section{Comparison with other general methods for addressing overfitting} \label{appenc}
There are some common experimental settings that may be used to reduce overfitting. For AdaLoRA, two promising methods are increasing weight decay and adopting a more aggressive rank pruning setting. We conducted experiments on these two methods separately, and the results indicate that neither of these approaches effectively addresses overfitting issues or enhances the model's generalization ability.

\subsection{Weight Decay}
We kept hyperparameters for AdaLoRA that have been well tuned in AdaLoRA and can achieve optimal results of AdaLoRA while only tuning the weight decay value. We conducted experiments on DeBERTa-v3-base on SST-2, CoLA and QNLI datasets. Results can be seen as follows.

\begin{table}[h] \centering
\caption{Experiment results on different weight decay values of AdaLoRA. }
\vspace{0.2cm}
\begin{tabular}{c|ccccccccc}
\toprule
 Weight Decay                    &SST-2      & CoLA   &QNLI   \\ \midrule
 0.00        & 95.80          & \textbf{70.04}   & \textbf{94.49}        \\ 
 0.05        & \textbf{95.96}  & 68.03         &  94.03                 \\ 
 0.10       &  95.30         & 68.42          &   93.52               \\ 
 0.20       &  94.60         & 67.87          &   93.76               \\ 
 \bottomrule
\end{tabular}
\label{table13}
\end{table}

\begin{figure}[h]
	
	\begin{minipage}{0.32\linewidth}
		\vspace{3pt}
		\centerline{\includegraphics[width=\textwidth]{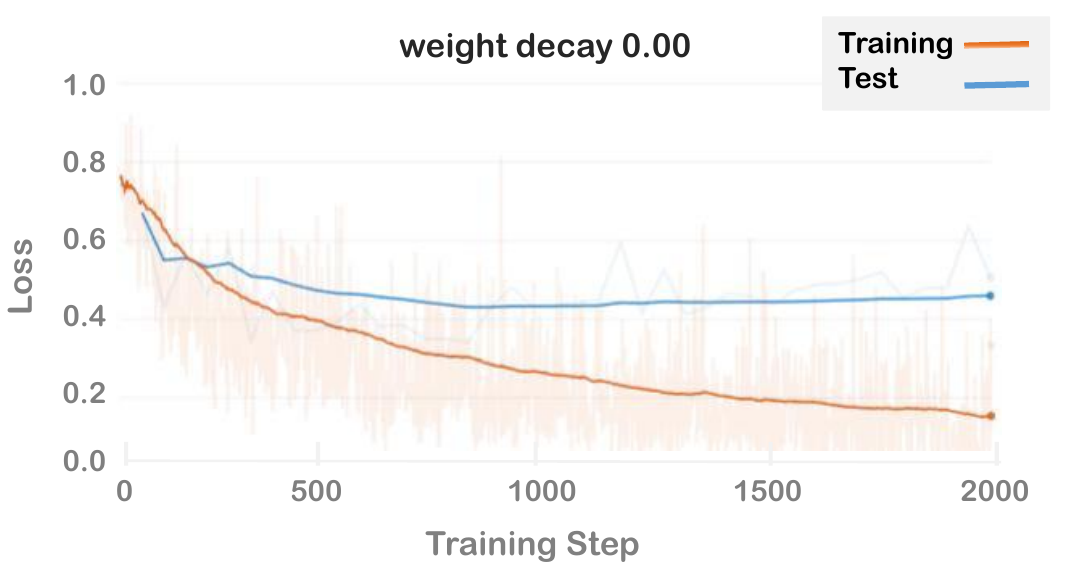}}
		\centerline{weight decay 0.00}
	\end{minipage}
	\begin{minipage}{0.32\linewidth}
		\vspace{3pt}
		\centerline{\includegraphics[width=\textwidth]{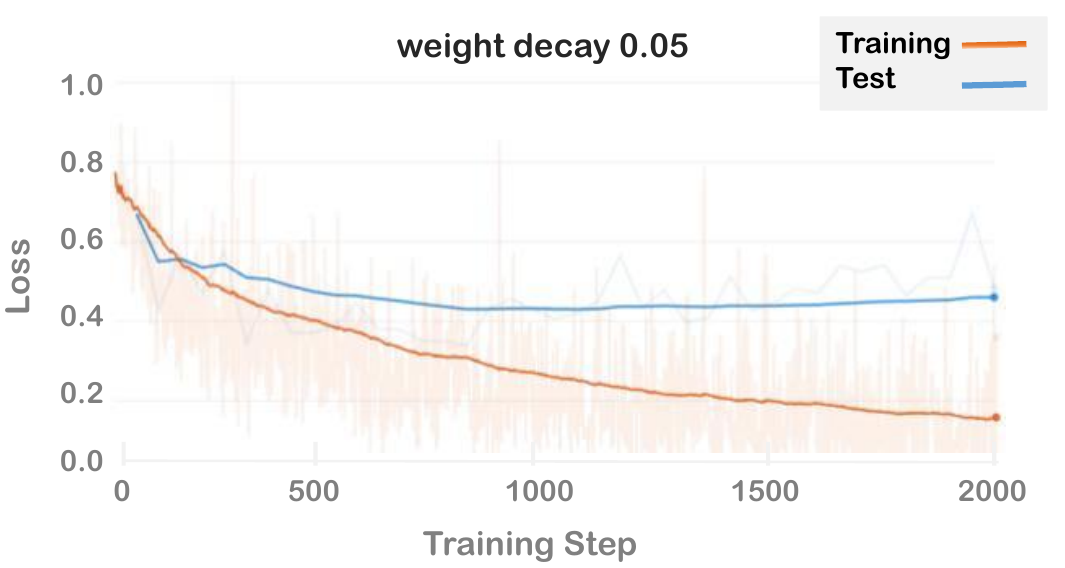}}
	 
		\centerline{weight decay 0.05}
	\end{minipage}
	\begin{minipage}{0.32\linewidth}
		\vspace{3pt}
		\centerline{\includegraphics[width=\textwidth]{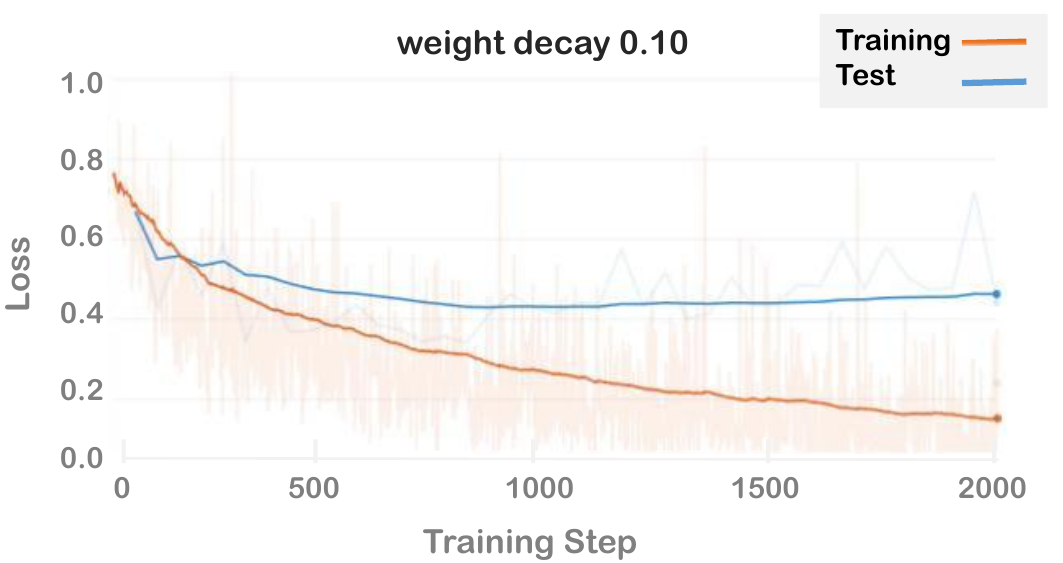}}
	 
		\centerline{weight decay 0.10}
	\end{minipage}
 
	\caption{Loss curves on CoLA training and test datasets for illustrating of the influence of different weight decays in AdaLoRA.}
\vspace{-0.4cm}
	\label{fig:abc1}
\end{figure}

In the context of consistent experimental configurations with other parameters, the application of an increased weight decay to AdaLoRA results in a decline in its performance. We further show the training/evaluation loss curves of different weight decay values for AdaLoRA on CoLA dataset in Figure~\ref{fig:abc1}. Divergent curves exhibit no obvious distinctions, and all display substantial gaps between training and evaluation losses. This suggests that increasing weight decay values in this situation does not effectively mitigate overfitting, consequently failing to enhance the model's generalization capability.

\subsection{Pruning rates of singular values}
During the fine-tuning process of AdaLoRA, the total rank budget is gradually decreasing to a target budget. Since the target rank for AdaLoRA is already one, we applied more aggressive pruning to rank budget through changing the LoRA-applied layer kind from six kinds to four kinds.
Specifically, in the experimental settings of AdaLoRA, we removed 'layer.output' and 'attention.output' as LoRA-applied layers and kept 'query', 'key', 'value', 'intermediate' as LoRA-applied layers. We conducted experiments on DeBERTa-v3-base on the CoLA dataset. Results can be seen as follows.

\begin{table}[h] \centering
\caption{Experiment results on different pruning rates of AdaLoRA. }
\vspace{0.2cm}
\begin{tabular}{c|ccccccccc}
\toprule
 Pruning Pattern            & CoLA     \\ \midrule
six kinds       & \textbf{70.04}                  \\ 
four kinds        & 67.97                \\ 
 \bottomrule
\end{tabular}
\label{table14}
\end{table}

\begin{figure}[t]
	
	\begin{minipage}{0.5\linewidth}
		\vspace{3pt}
		\centerline{\includegraphics[width=\textwidth]{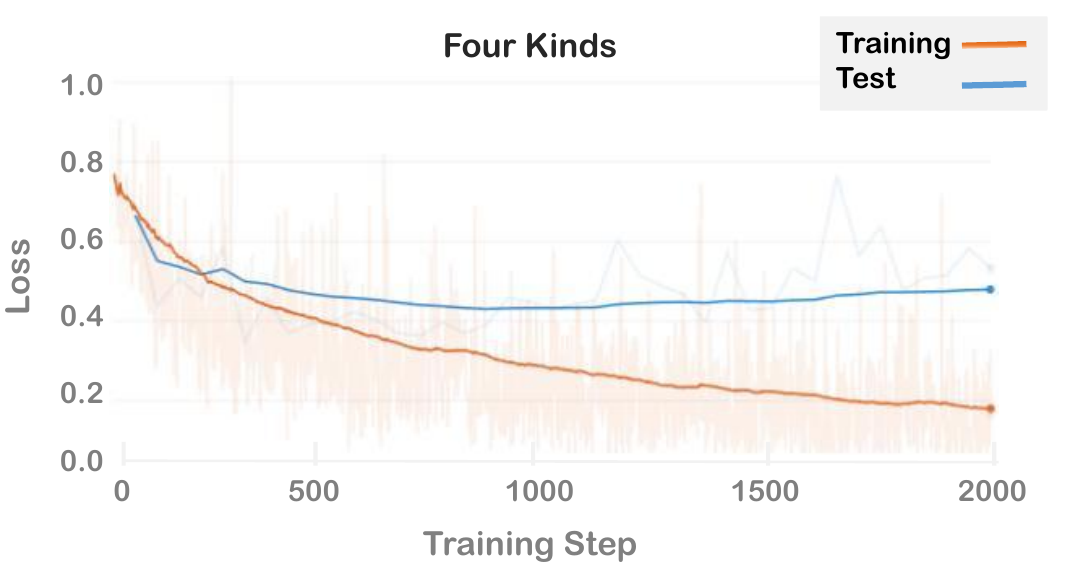}}
	\end{minipage}
	\begin{minipage}{0.5\linewidth}
		\vspace{3pt}
		\centerline{\includegraphics[width=\textwidth]{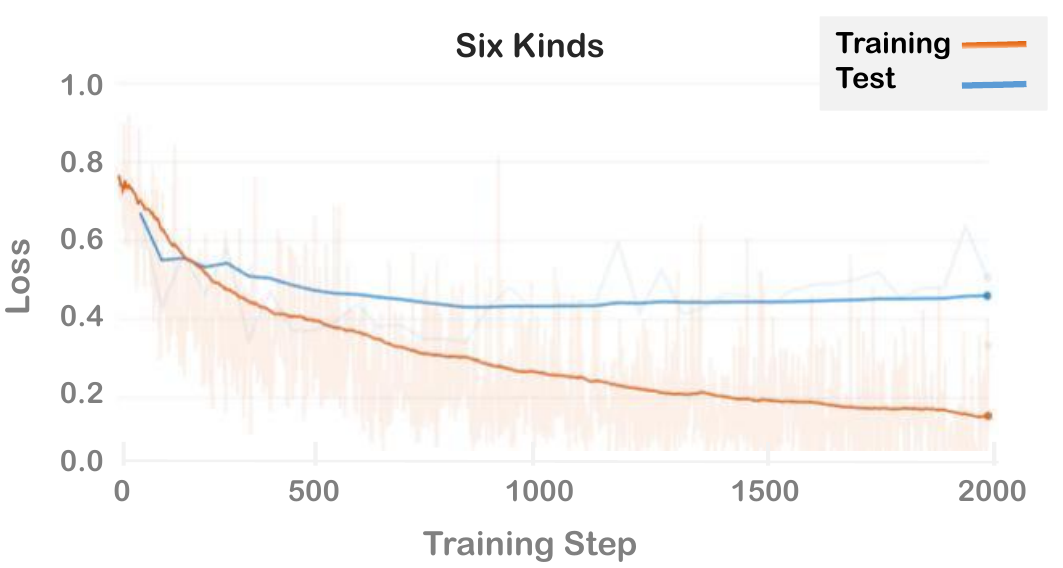}}
	 
	\end{minipage}
 
	\caption{Loss curves on CoLA training and test datasets for illustrating of the influence of different pruning settings in AdaLoRA.}
\vspace{-0.4cm}
	\label{fig:ab2}
\end{figure}

In the context of consistent experimental configurations with other parameters, the application of more aggressive pruning rates to AdaLoRA results in a decline in its performance. We further show the training/evaluation loss curves of two pruning rates for AdaLoRA on CoLA dataset in Figure~\ref{fig:ab2}. Divergent curves exhibit no obvious distinctions, and all display substantial gaps between training and evaluation losses. This suggests that a more aggressive pruning strategy in this situation does not effectively mitigate overfitting, consequently failing to enhance the model's generalization capability.

\section{The distribution of fine-tuned singular values}\label{append}

\begin{figure}[h]
	
	\begin{minipage}{0.5\linewidth}
		\vspace{3pt}
		\centerline{\includegraphics[width=\textwidth]{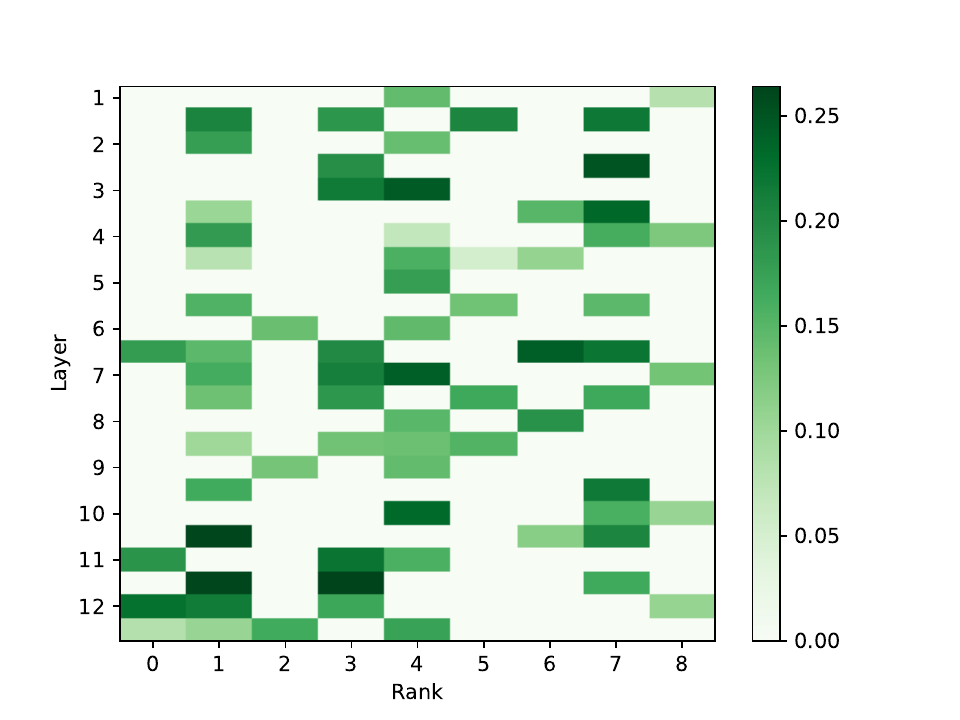}}
		\centerline{AdaLoRA Singular Value Distribution}
	\end{minipage}
	\begin{minipage}{0.5\linewidth}
		\vspace{3pt}
		\centerline{\includegraphics[width=\textwidth]{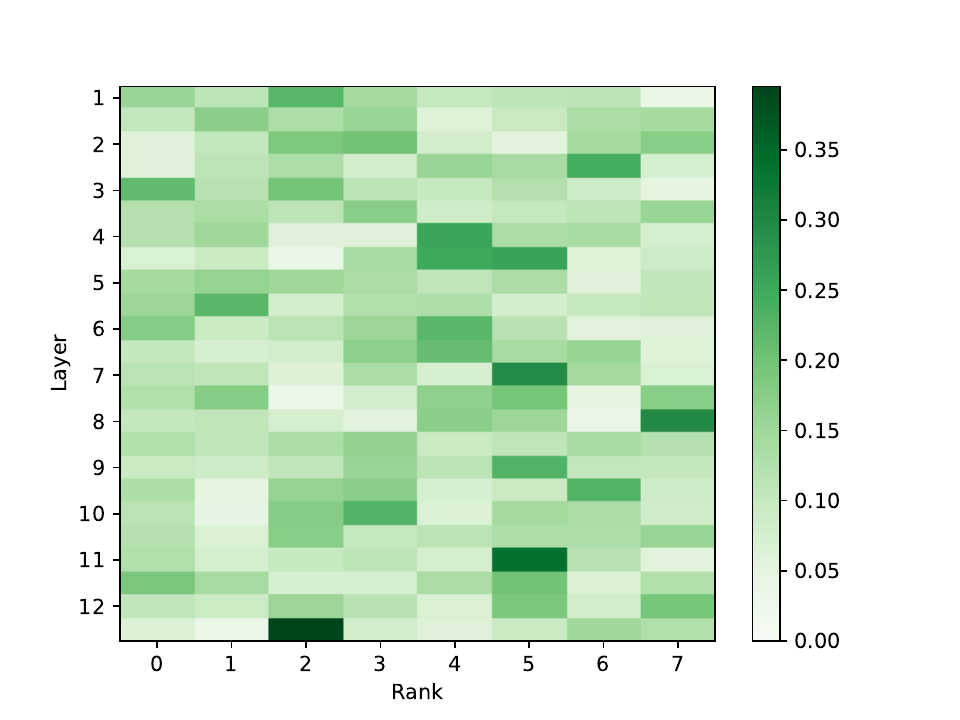}}
	 
		\centerline{BiLoRA Singular Value Distribution}
	\end{minipage}
 
	\caption{Singular Value Distribution of BiLoRA and AdaLoRA.}
\vspace{-0.4cm}
	\label{fig:ab3}
\end{figure}

The distribution of singular values serves as an indicator of the contributions of distinct singular vectors to the corresponding incremental matrix.
We plotted the distribution of singular values learned both in BiLoRA and AdaLoRA in Figure\ref{fig:ab3}. We used the singular values of RoBERTa-base on the CoLA dataset. AdaLoRA implements low-rank adapters to all six linear layers while BiLoRA only implements low-rank adapters to 'query' and 'key' layers.

The distribution of singular values in AdaLoRA and BiLoRA share various similar features: 1) Several singular values are nearly zero. AdaLoRA gradually prunes ranks to make some of the singular values zero. BiLoRA utilizes bi-level optimization to automatically learn this distribution. 2) A small number of highly significant singular values have yielded considerable contributions.

More results can be analyzed in future work in the distribution of singular values of optimal solutions. We expect BiLoRA can provide flexible interaction with singular values through separately tuning the upper-level training strategy and help us better understand the process of low-rank fine-tuning.


\section{The orthogonality of singular vectors}\label{appene}
Ablation study on orthogonality regularization for the singular vectors illustrates that the performance is largely unaffected by the coefficient of this regularization. We further investigate the reason for the robustness of BiLoRA against orthogonal regularization. 
With the $\Delta W = P \Lambda Q$ parameterization, we plotted the value distribution of $P^{T}P$ for a random $P$ to illustrate the orthogonality of singular vectors just after initialization, with regularization and with $\lambda = 0.1$ regularization.

\begin{figure}[h]
	
	\begin{minipage}{0.32\linewidth}
		\vspace{3pt}
		\centerline{\includegraphics[width=\textwidth]{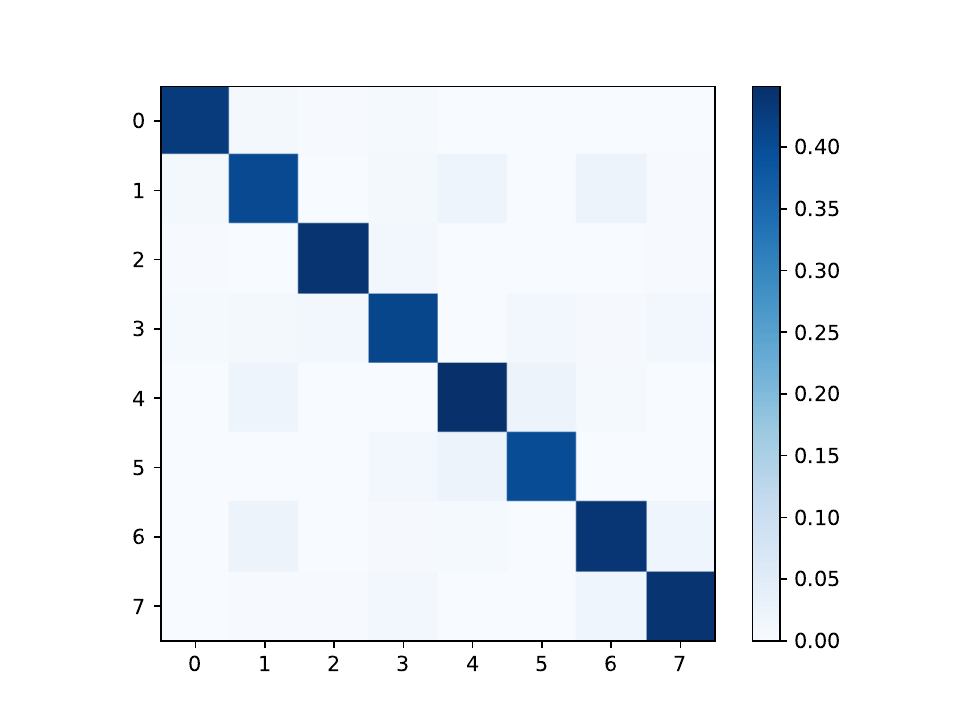}}
		\centerline{Without Regularization}
	\end{minipage}
	\begin{minipage}{0.32\linewidth}
		\vspace{3pt}
		\centerline{\includegraphics[width=\textwidth]{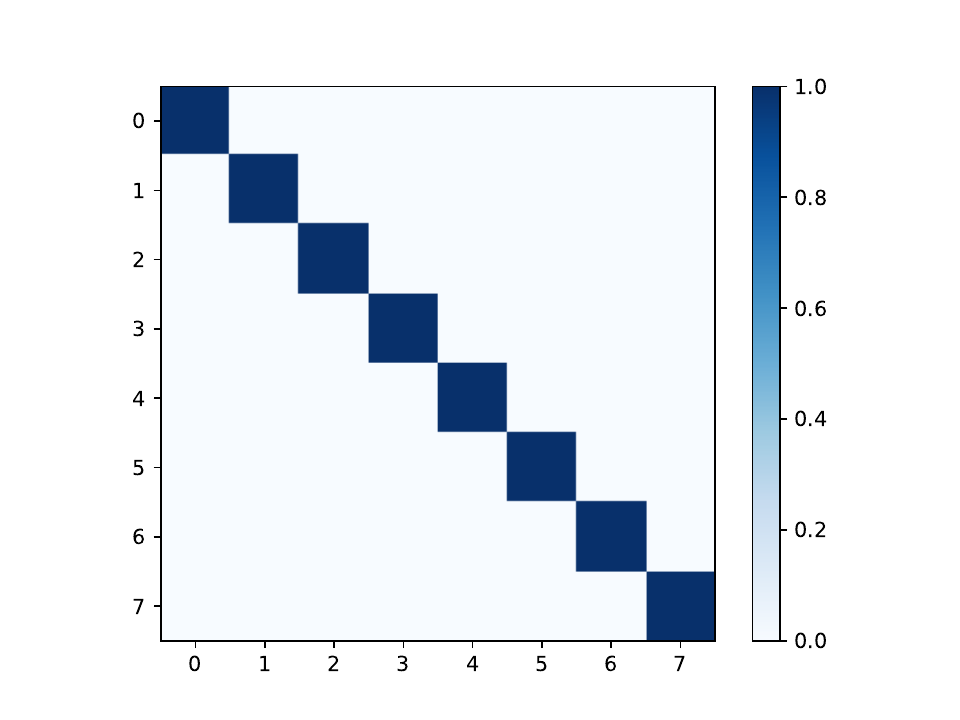}}
	 
		\centerline{Regularization with $\lambda=0.1$ }
	\end{minipage}
	\begin{minipage}{0.32\linewidth}
		\vspace{3pt}
		\centerline{\includegraphics[width=\textwidth]{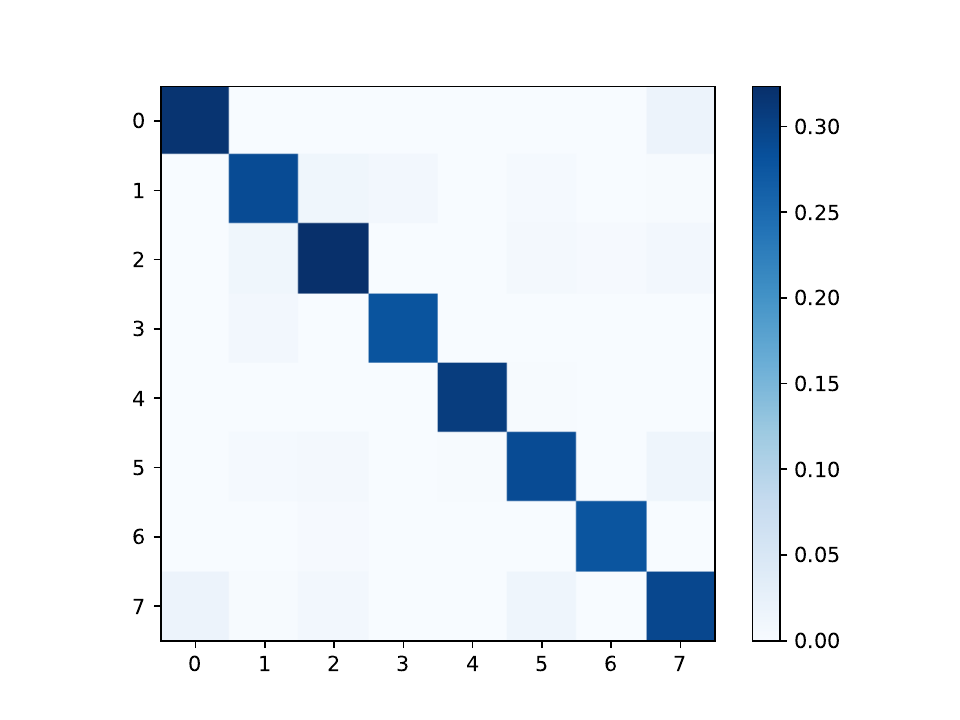}}
	 
		\centerline{Initialization}
	\end{minipage}
 
	\caption{The value distribution of singular vectors, relecting the orthogonality of singular vectors.}
\vspace{-0.4cm}
	\label{fig:ab4}
\end{figure}

\begin{figure}[H] 
\centering 
        \begin{minipage}{0.32\linewidth}
		\vspace{3pt}
		\centerline{\includegraphics[width=\textwidth]{lora.pdf}}
		\centerline{LoRA}
	\end{minipage}
        \begin{minipage}{0.32\linewidth}
		\vspace{3pt}
		\centerline{\includegraphics[width=\textwidth]{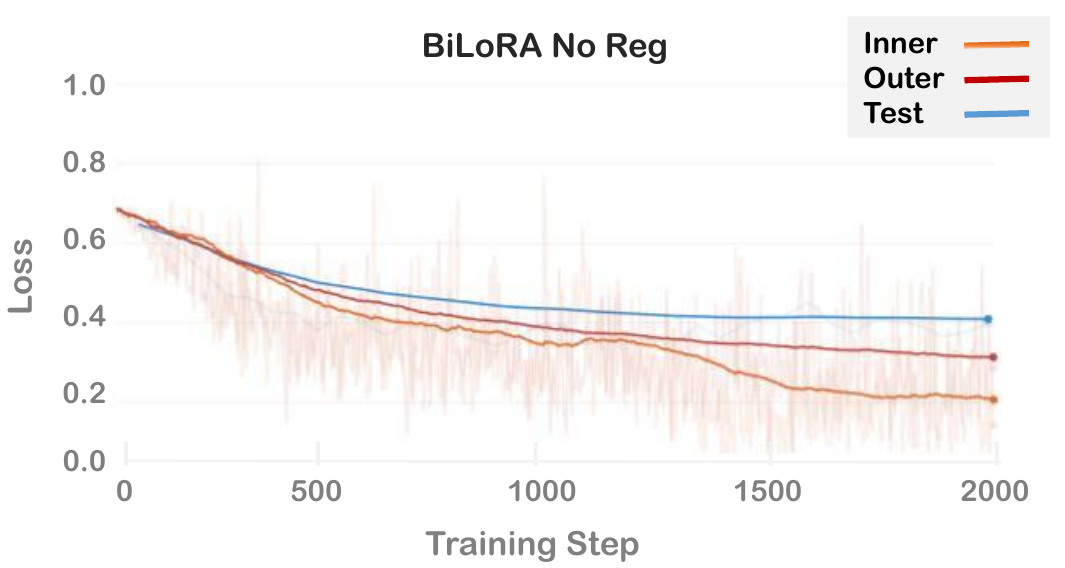}}
		\centerline{BiLoRA without Reg}
	\end{minipage}
	\begin{minipage}{0.32\linewidth}
		\vspace{3pt}
		\centerline{\includegraphics[width=\textwidth]{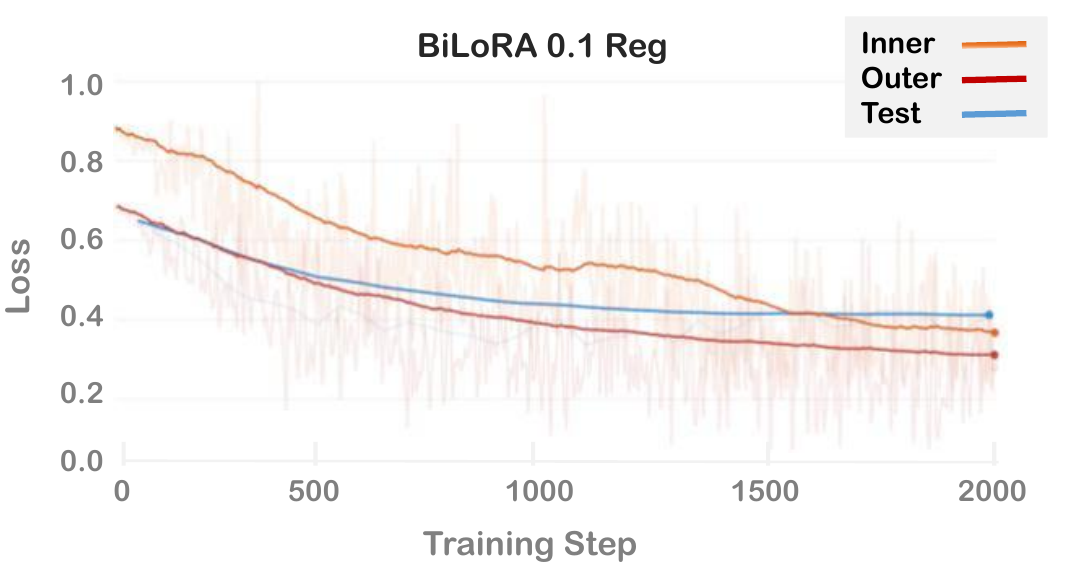}}
	 
		\centerline{BiLoRA with $\lambda=0.1$ Reg }
	\end{minipage}
 
\caption{Loss curves on CoLA training and test datasets for illustrating of the influence of regularization in BiLoRA. }
\label{fig:final} 
\end{figure}

From Figure~\ref{fig:ab4}, we suspect the reason for the robustness of BiLoRA against orthogonal regularization can be from these 3 aspects:
\begin{itemize}
    \item The singular vector matrices inherit a degree of “natural orthogonality" without regularization just after initialization due to the 'Normal Initialization' we use for initializing all the singular vector matrices.
    \item The value distribution of $P^{T}P$ and $QQ^{T}$ in optimal solutions without regularization is close to Identity Matrix. Despite magnitudes, singular vectors are largely orthogonal. When regularization coefficient $\lambda \geq 0.1$, singular matrices are almost orthogonal after warmup. 
    \item Optimal solutions with different regularization coefficients including $\lambda = 0.0$ can all contribute greatly to mitigating overfitting. This is illustrated in Figure~\ref{fig:final}.
\end{itemize}

\section{The role of various hyper-parameters in BiLoRA }\label{appenf}
The hyperparameter tuning for BiLoRA is quite simple, convenient and straightforward since BiLoRA only introduces an additional level of learning rate compared to LoRA and has fewer hyperparameters than AdaLoRA. We further conducted experiments with regard to the dataset partition of $D_1 $ and $ D_2$ and the unroll steps $T_1 $ and $ T_2$ to offer insights of their role in BiLoRA.

\textbf{Data Partition of $D_1 $ and $ D_2$.} The Dataset Partition, together with learning rate can help keep the balance of inner/outer optimization, which can contribute to preventing the model from overfitting. Lower level has more trainable parameters, so it’s natural to use more data for training singular vectors, while using the left for training singular values.
We experimented on DeBERTa-v3-base on CoLA and SST-2 datasets to show the influence of different dataset partitions. We change the inner level dataset $D_1$ partition from $0.6$ to $1.0$ with $0.1$ interval. The results can be found in Table~\ref{table15}.

\begin{table}[h] \centering
\caption{Experiment results on different data partitions of BiLoRA. }
\vspace{0.2cm}
\begin{tabular}{c|ccccccccc}
\toprule
 Partition            & CoLA  & SST-2  \\ \midrule
0.6       & 68.01	&94.84             \\ 
0.7        & 70.16	&95.87              \\ 
0.8        & \textbf{70.52}	&96.02               \\ 
0.9        & 70.39	&\textbf{96.44}             \\ 
1.0        & 67.94	&94.51    \\ 
 \bottomrule
\end{tabular}
\label{table15}
\vspace{-0.3cm}
\end{table}

Results show that too small partitions ($<=0.6$)  or too large partitions ($1.0$, only singular vectors are trained) can harm the overall performance. When the inner partition is too small, singular vectors are not well trained and when the inner partition is 1.0, singular vectors are not trained at all, which can cause a great performance drop. These results can also demonstrate that the bi-level optimization is efficient and the two levels are both necessary in preventing overfitting and enhancing performances. In the paper, we don’t ever change the partition of the data and keep it $8:2$. Tuning the partition may further improve the overall performance.

\textbf{Unroll Steps $T_1$ and $T_2$.} In previous experiment results, we didn’t tune the Unroll Steps throughout our experiments and kept $T_1 = T_2$. Actually there can be a set of hyperparameters that can all achieve good results because they can all keep good balance of inner/outer level from 3 perspectives: 1) performing well on inner dataset; 2) performing well on outer dataset; 3) not overfitting on either subset and generalizing well on test dataset. We further conducted experiments on on DeBERTa-v3-base on CoLA dataset with different Unroll Steps. All other hyperparameters are kept the same.

\vspace{-0.3cm}
\begin{table}[h] \centering
\caption{Experiment results on different unroll steps of BiLoRA. }
\vspace{0.2cm}
\begin{tabular}{cc|ccccccccc}
\toprule
 $T_1$ &   $T_2$      & CoLA    \\ \midrule
1       & 1	 &\textbf{70.52}         \\ 
1       & 3	  &	70.40              \\ 
2       & 1	     &69.96	             \\ 
2        & 3	&70.14             \\ 
5       & 1	   &70.01	    \\ 
5       & 3	  &69.19    \\ 
 \bottomrule
\end{tabular}
\label{table16}
\vspace{-0.3cm}
\end{table}

The total optimization steps for inner level and outer level are close for different $T_1/T_2$. Typically, a single inner optimization step is faster than a single outer optimization step due to the calculation of hypergradients of outer level. So using a larger $T_1$ is also an efficient and practical choice.

We don’t exactly tune the iteration numbers for 2 reasons.
\begin{itemize}
    \item $T_1 = T_2 = 1$ is an empirical choice in existing bi-level optimization tasks. The main effect of unroll steps can be the balance of inner and outer level. Intuitively and practically, $T_1 = T_2 = 1$ means BiLoRA frequently optimizes between the two levels, preventing the model from overfitting on either subset and effectively addresses overfitting.

    \item We expect BiLoRA to be powerful and effective, yet easy to use. The number of hyperparameters for BiLoRA is kept nearly the same as LoRA, which is much less than AdaLoRA.
\end{itemize}

\section{Additional comparison with LoRA and AdaLoRA}\label{appeng}

For a more thorough comparison among BiLoRA, LoRA and AdaLoRA in both performances and computation costs, we further conducted experiments on RoBERTa-base on 4 NLU tasks and on GPT-2 on 2 NLG tasks. The experiment results can be seen as below, higher is better for all the scores. 

\subsection{Performances}
First, we compared BiLoRA with AdaLoRA on 4 NLU datasets. The resuls are shown in Table \ref{table17}. Experiment results show that BiLoRA surpasses AdaLoRA on all the 4 datasets with a notable performance gap on average, further demonstrating the effectiveness of BiLoRA.

Second, we compared BiLoRA with AdaLoRA, LoRA and other baselines on 2 more NLG datasets, WebNLG and DART in Table \ref{table18}. We reported the BLEU score and higher is better. Results of other methods are from LoRA as a score reference.

Results on different models and on both NLU and NLG datasets show that our method, BiLoRA outperforms LoRA, AdaLoRA and other baselines by a large margin, demonstrating the effectiveness of our method.

\vspace{-0.2cm}
\begin{table}[t] \centering
\caption{Experiment results with regard to BiLoRA and AdaLoRA on 4 NLU datasets.}
\vspace{0.2cm}
\begin{tabular}{c|cccccccccc}
\toprule
Method  &MNLI   & CoLA  &QNLI   & QQP  & Avg.    \\ \midrule
AdaLoRA      & 87.2	 &63.4	&92.8	&91.0	&83.6         \\ 
BiLoRA       & \textbf{87.9}	 &\textbf{64.8}	&\textbf{93.3}	&\textbf{91.4}	&\textbf{84.4}              \\ 
 \bottomrule
\end{tabular}
\label{table17}
\vspace{-0.3cm}
\end{table}

\vspace{-0.2cm}
\begin{table}[h] \centering
\caption{Experiment results with regard to BiLoRA, LoRA, AdaLoRA and other baselines on 2 more NLG datasets, WebNLG and DART.}
\vspace{0.2cm}
\begin{tabular}{l|ccccccccc}
\toprule
 Method   &  WebNLG	      & DART    \\ \midrule
Full Fine-tune	&46.5	&46.2 \\
Adapter\textsuperscript{L}	&50.2	&42.4 \\
FT\textsuperscript{top2}	&54.9	&41.0\\
Prefix	&36.0	&46.4\\
LoRA	&55.3	&47.1\\
AdaLoRA	&55.5	&47.6\\
BiLoRA	&\textbf{56.1}	&\textbf{49.0} \\
 \bottomrule
\end{tabular}
\label{table18}
\vspace{-0.3cm}
\end{table}

\subsection{Computation Costs.}
Firstly, we provided the total training steps needed for convergence for LoRA and BiLoRA and the per-update cost for the two methods in Table \ref{table19}. We use the results of RoBERTa-base on MNLI and SST-2 datasets. The per-update cost is measured in $min/k$.

\vspace{-0.3cm}
\begin{table}[h] \centering
\caption{Experiment results with regard to BiLoRA and LoRA on computation costs.}
\vspace{0.2cm}
\begin{tabular}{ll|cccccccc}
\toprule
  &Method   &  MNLI	      & SST-2    \\ \midrule
Total steps  & RoBERTa(LoRA)  &	184k &	63k \\
  & RoBERTa(BiLoRA) &	2$\times$15k(1/6$\times$ )	& 2$\times$3k(1/11$\times$ )\\ \midrule
Per-step cost & RoBERTa(LoRA)&	17.34	 &17.40  \\
 & RoBERTa(BiLoRA)&	46.90(2.7$\times$ ) &	43.33(2.49$\times$ ) \\
 \bottomrule
\end{tabular}
\label{table19}
\end{table}

\vspace{-0.2cm}
\begin{table}[H] \centering
\caption{Experiment results with regard to BiLoRA and AdaLoRA on computation costs.}
\vspace{0.2cm}
\begin{tabular}{ll|cccccccc}
\toprule
  &Method   &  MNLI	      & SST-2    \\ \midrule
  Total time &Dv3(AdaLoRA)	&753.54	 &240.57 \\
  &Dv3(BiLoRA)	&446.21(1/1.7$\times$)	 &56.71(1/4.2$\times$) \\ \midrule
Total steps  & Dv3(AdaLoRA)	&85.9k	&50.5k \\
& Dv3(BiLoRA)	&2$\times$15k(1/3$\times$)	&2$\times$3k(1/8$\times$) \\ \midrule
Per-step cost & Dv3(AdaLoRA)	&8.77	&4.76  \\
&Dv3(BiLoRA)	&14.87(1.7$\times$)	&9.45(2.0$\times$) \\
 \bottomrule
\end{tabular}
\label{table20}
\vspace{-0.3cm}
\end{table}

Results show that BiLoRA uses 6 times/11 times less steps than LoRA for convergence. The per-step cost for BiLoRA is roughly 2.5 times as LoRA, since BiLoRA needs to iteratively optimize between the two levels and the calculation of outer hypergradients can cost more than simple gradient calculation. Results demonstrate that BiLoRA can converge much faster than LoRA and thus takes much less time for training than LoRA.

Secondly, we provide the total training steps needed for convergence for AdaLoRA and BiLoRA and the per-update cost for the two methods in Table \ref{table20}. We use the results of DeBERTa-v3-base on MNLI and SST-2 datasets. According to AdaLoRA, each training epoch of AdaLoRA is longer than LoRA, thus intuitively BiLoRA is also faster than AdaLoRA. The per-update cost is measured in $min/k$ and time is measured in $min$.

Results show that BiLoRA uses 3 times/8 times less steps than AdaLoRA for convergence. The per-step cost for BiLoRA is roughly 1.7/2.0 times as AdaLoRA. Results demonstrate that BiLoRA can converge much faster than LoRA, AdaLoRA and take much less time for training than baselines.

\end{document}